\documentclass[subscriptcorrection,upint,varvw,barcolor=Goldenrod3,mathalfa=cal=euler,balance,hyphenate,french,pdf-a]{asmejour} %

\usepackage{tabularx}
\usepackage{multicol}
\usepackage{multirow}
\usepackage{placeins}
\usepackage{cleveref}
\usepackage{caption}    
\crefname{appendix}{Appendix}{Appendices}
\crefname{table}{Table}{Tables}
\crefname{figure}{Figure}{Figures}
\usepackage{times}

\newcolumntype{Y}{>{\centering\arraybackslash}X}
\newcolumntype{Z}{>{\centering\arraybackslash}p{2cm}}

\hypersetup{%
	pdfauthor={John H. Lienhard},                       		   	
	pdftitle={ASME Journal Paper LaTeX Template},                  	
	pdfsubject = {Describes the asmejour LaTeX template},			
	pdflicenseurl={https://ctan.org/pkg/asmejour},
}

\begin{document}


\SetAuthorBlock{Phillip Mueller\CorrespondingAuthor}{BMW Group,\\
Knorrstrasse 147,\\
80788 Munich, Germany,\\
email: phillip.mueller@bmw.de} 

\SetAuthorBlock{Lars Mikelsons}{
Chair for Mechatronics, \\
Department of Applied Computer Science,\\
University of Augsburg, \\
Am Technologiezentrum 8, 86159 Augsburg, Germany \\
email: lars.mikelsons@uni-a.de
}

\title{Exploring the Potentials and Challenges of Deep Generative Models in Product Design Conception}


\begin{abstract}
The synthesis of product design concepts stands at the crux of early-phase development processes for technical products, traditionally posing an intricate interdisciplinary challenge. The application of deep learning methods, particularly Deep Generative Models (DGMs), holds the promise of automating and streamlining manual iterations and therefore introducing heightened levels of innovation and efficiency. However, DGMs have yet to be widely adopted into the synthesis of product design concepts. This paper aims to explore the reasons behind this limited application and derive the requirements for successful integration of these technologies. We systematically analyze DGM-families (VAE, GAN, Diffusion, Transformer, Radiance Field), assessing their strengths, weaknesses, and general applicability for product design conception. 
Our objective is to provide insights that simplify the decision-making process for engineers, helping them determine which method might be most effective for their specific challenges. Recognizing the rapid evolution of this field, we hope that our analysis contributes to a fundamental understanding and guides practitioners towards the most promising approaches. This work seeks not only to illuminate current challenges but also to propose potential solutions, thereby offering a clear roadmap for leveraging DGMs in the realm of product design conception.

\end{abstract}

\date{Version \versionno, \today}

\maketitle 


\section{Introduction}

Product design conception (PDC) is an intricate and multifaceted process, demanding considerable effort and investment. During the early-phase, foundational design concepts translate functional and feature requirements into preliminary visual representations \cite{ulrichProductDesignDevelopment2008}. These visualizations are crucial, shaping both the aesthetic appeal and engineering functionality of the final product. The challenge intensifies in consumer products such as passenger vehicles, which must harmonize appealing design with robust functionality to meet both aesthetic desires and practical necessities.

Traditionally, this process heavily relies on the domain expertise of engineers and designers, necessitating extensive manual iteration across various design modalities. Such iterative and manual processes are not only time-consuming but also require substantial investments, often depending on limited and compartmentalized information that impede innovation and efficiency. 
Recent advances in deep learning, particularly in generative methods for image synthesis, offer a promising path to the goal of reducing manual, time-consuming iterations. Additionally, they bear the potential to elevate human creativity by democratizing the engineering design process through lowering the skill-barrier \cite{millerHowShouldWe2021}.

Deep Generative Models (DGMs) are specifically tailored to learn complex data distributions and generate novel samples from the learned distributions \cite{kingmaAutoEncodingVariationalBayes2013,sohl-dicksteinDeepUnsupervisedLearning2015}. These models, including Generative Adversarial Networks (GANs), Variational Autoencoders (VAEs), Diffusion Models, and Transformer-based architectures, have demonstrated significant potential in various fields, including natural language processing \cite{vaswaniAttentionAllYou2017} and image generation \cite{hoDenoisingDiffusionProbabilistic2020,songDenoisingDiffusionImplicit2022,rombachHighResolutionImageSynthesis2022,podellSDXLImprovingLatent2023,sahariaPhotorealisticTexttoImageDiffusion2022}. 
The landscape of DGM-applications is currently dominated by text-based tasks and the corresponding Large-Language-Models like LLAMA \cite{touvronLlamaOpenFoundation2023} and GPT \cite{openaiChatGPT2023}.

Despite their undeniable potential, DGMs for visual data generation have not yet been widely adopted in engineering design and more specifically in PDC \cite{2024SteiningerPotentialsProductDesign}. This is due to several challenges that complicate their integration into existing workflows. Firstly, product and engineering design tasks often require precise domain-specific knowledge, which is difficult to encapsulate within the frameworks of DGMs. These tasks rely on specific modalities, like requirement tables, hand-drawn sketches, technical drawings and low-fidelity images to translate ideas and constraints into technical product representations. Many of the modalities are challenging for DGMs to accommodate. As a result, the outputs from DGMs frequently lack the robustness, reliability, interpretability, and replicability necessary for critical design applications \cite{alamAutomationAugmentationRedefining2024,joskowiczEngineersPerspectivesUse2023}.

Secondly, the inherent complexity of DGMs poses a significant barrier to their adoption by non-expert users, such as engineers, who may find the advanced machine learning concepts and operations daunting. This challenge is compounded by the rapid evolution and diversity of models within the field, which can overwhelm even dedicated specialists trying to keep pace with the latest developments \cite{maslejAIIndex20242024}. Selecting a feasible method for a given task typically requires elaborate experimentation on top of existing in-depth knowledge by an expert.
Additionally, efficient application and deployment of DGMs within the PDC-process presuppose substantial data and computational resources, further restricting their accessibility and scalability. These models must also effectively integrate multifaceted data that captures human inputs and interactions, a requirement essential for enhancing design quality and tailoring solutions to specific domain challenges. The convergence of these factors contributes to the limited penetration of DGMs in the realm of product design conception.

This study aims to provide a comprehensive analysis of the potentials and challenges associated with the application of visual DGMs in PDC. We specifically focus on visual (2D) representations, including both simple parametric shapes and higher-fidelity images of product concepts. These modalities are particularly relevant in early-phase concept development, where rapid, low-fidelity sketches and 2D silhouettes guide brainstorming and enable quick iterations \cite{chenAerodynamicDesignOptimization2019, chenInverseDesignTwoDimensional2022}. In an effort to lower the barrier for the application of these models, we will derive fundamental requirements that DGMs must fulfill to be effectively integrated into PDC-processes. These requirements will serve as a basis for evaluating important model families, assessing their suitability for tasks within product design. By examining how these approaches align with the defined requirements, we seek to enhance understanding and assist practitioners in three key areas:
\begin{enumerate}
    \item Analyzing Model Suitability: Practitioners will gain a better and more realistic understanding of which applications are realistic with DGMs and where their limitations lie. This will facilitate more informed decisions regarding the selection of appropriate models for solving specific product design problems.
    \item Evaluating Application Cases: By outlining prerequisites and limitations of DGMs concerning PDC-application, we hope to guide practitioners in analyzing their application cases to determine the suitability for incorporating DGMs. This analysis aims to ensure that the deployment of these technologies is both practical and effective, taking into consideration the specific conditions and constraints of the design task at hand.
    \item Outlining Research Directions: As a result of our analysis, we aim to identify areas where further research could enhance the applicability and effectiveness of DGMs. 
\end{enumerate}

By setting a baseline for the comparison of DGM-capabilities with the needs of the concept design process, we hope to assist in selecting the most appropriate models, setting realistic expectations, and effectively utilizing these technologies in product design workflows. 

We summarize relevant background about DGM-families in \Cref{sec:BG} and provide further references for the interested reader. \Cref{sec:DGM_PDC} discusses the PDC-process, existing potentials and open challenges, from which the key requirements for DGM-application are derived. The outlined requirements form the basis for the technical analysis of DGM-families in \Cref{sec:Analysis}. To lower the knowledge-barrier and make our work more accessible for non-DGM-experts, we summarize our application recommendations in \Cref{sec:recommend}. Our conclusions in \Cref{sec:Conlusion} aim to briefly discuss the three objectives we outlined throughout this chapter. We provide additional examples in the appendix (\Cref{sec:Appendix}).

\section{Background}
\label{sec:BG}
\subsection{Deep Generative Models}

\subsubsection{Generative Adversarial Networks (GANs)} GANs are a class of models that consist of two competing neural networks – the generator and the discriminator. GANs have found success in a variety of applications since their initial introduction \cite{goodfellowGenerativeAdversarialNets}, including image generation \cite{karrasStyleBasedGeneratorArchitecture2021,karrasAnalyzingImprovingImage2020} and manipulation \cite{isolaImagetoImageTranslationConditional2017}, as well as text-to-image synthesis \cite{sauerStyleGANTUnlockingPower2023}. While the generator learns to map a vector of latent variables to a desired distribution, thus generating new content, the discriminator learns to distinguish between real and generated content. The weights of both networks are improved independently during training. As the discriminator improves, the generator also improves as it learns to generate content that fools the discriminator. The training of GANs is often considered challenging due to instabilities from vanishing gradients in the generator and the failure of the generator to capture all modes in the data distribution (mode collapse). Nevertheless, GAN-architectures have found great success in a variety of applications, as they allow for the important possibility of conditioning the generative process, meaning the generative process can be guided by user-provided constraints. This significantly increases the potential for application in product design. Reference \cite{mirzaConditionalGenerativeAdversarial2014} provide a discretely conditioned GAN (‘cGAN’), where the conditioning vector is fed into the generator and the discriminator. There are numerous other approaches on conditioning GANs, we direct the interested reader to refer to ‘InfoGAN’ \cite{chenInfoGANInterpretableRepresentation2016} and continuous conditional GANs \cite{dingCCGANCONTINUOUSCONDITIONAL2021}.

\subsubsection{Variational Autoencoder (VAEs)} VAEs belong to the family of probabilistic machine learning models. For Autoencoders, an encoder maps the input data into a lower-dimensional latent representation while the decoder reconstructs the original content as accurately as possible. Variational Autoencoders, introduced by Kingma and Welling \cite{kingmaAutoEncodingVariationalBayes2013}, add probabilistic sampling in the latent space, regularizing the latent distribution and creating a more continuous mapping of the data distribution to the latent space that allows for sampling of realistic latent vectors. The Kullback-Leibler (KL) divergence loss between the latent distribution and a standard Gaussian is added to obtain a predictable latent space distribution. Interested readers may refer to \cite{kingmaIntroductionVariationalAutoencoders2019,princeUnderstandingDeepLearning2023} for detailed information on VAE-foundations and to \cite{murphyProbabilisticMachineLearning2023,sohnLearningStructuredOutput2015} for conditional VAEs.

\subsubsection{Diffusion models} Diffusion-based architectures are a class of generative models that simulate the data generation as a reverse diffusion process. The approach, routed in non-equilibrium thermodynamics, was initially proposed in 2015 \cite{sohl-dicksteinDeepUnsupervisedLearning2015}. Data structures, like images, are gradually destroyed through the addition of Gaussian noise until there is no information left from the original data. The reverse process is learned by a deep neural network, that transforms a simple distribution (like Gaussian noise) into the complex distribution of the original data. Although requiring large amounts of data and compute, diffusion models have recently found immense success in image generation. We direct the interested reader to the following works: \cite{hoDenoisingDiffusionProbabilistic2020,nichol2021improveddenoisingdiffusionprobabilistic,dhariwalDiffusionModelsBeat2021,songDenoisingDiffusionImplicit2022} for fundamentals and \cite{rombachHighResolutionImageSynthesis2022} as well as \cite{sahariaPhotorealisticTexttoImageDiffusion2022} for large-scale image generation applications. 

\subsubsection{Transformers} The Transformer is an autoregressive deep learning architecture popular in the fields of natural language processing and computer vision. The architecture was initially proposed in the widely known work ‘Attention Is All You Need’ \cite{vaswaniAttentionAllYou2017}, to which we refer the reader for more detailed information. Transformers rely on the ‘attention mechanism’, that computes a weighted sum of input values (V) based on their ‘attention scores’ (relevance) to a specific query (Q) and key (K) pair. This allows to capture global dependencies between input and output in parallel and dynamically prioritize specific parts of the input data. Transformers excel at capturing long-range dependencies and intricate patterns in data. Transformers have been widely adopted for training large language models \cite{devlinBERTPretrainingDeep2019,brownLanguageModelsAre2020,radfordLanguageModelsAre2019}. They also are a key element in vision-centric models, to learn visual concepts from natural language, such as in ‘ViT’ \cite{dosovitskiyImageWorth16x162021} and ‘CLIP’ \cite{radfordLearningTransferableVisual2021}. Noteable works for visual data generation include \cite{esserTamingTransformersHighResolution2021, esser2021imagebart, chang2022maskgit, peeblesScalableDiffusionModels2023}. Coupled with a flow matching loss, the Transformer architecture forms the basis for recently developed image generation models that surpass the generative capabilities of Diffusion-based approaches \cite{esserScalingRectifiedFlow2024,flux2024}.

\subsection{Development of Early-Phase Product Design Concepts}
\subsubsection{Process}
The early-phase process of PDC is characterized by the definition and subsequent translation of main engineering requirements into functional representations of the product design. In the product-design-process, the PDC is localized in the early-phases of “Ideation and Planning” as well as “Concept Development” \cite{ulrichProductDesignDevelopment2008}. The phase of planning and ideation is crucial for brainstorming and comparing diverse ideas, facilitating a creative exploration of potential solutions before formalizing the product concept. It serves as the foundation for generating a broad array of ideas, which are then critically evaluated to translate them into an initial set of requirements for the product and its design. It is essential for exploring the realm of physical and technical possibilities, identifying domain-specific prerequisites, and setting the stage for defining a core set of product requirements. As the process transitions into subsequent conceptualization in Phase 1, these initial ideas are further refined.

Transitioning into “Concept Development”, the focus shifts to the design conception, where the visualization and iteration of product design concepts become paramount. The objective is to converge opposing technical and design demands into a meaningful design. This requires numerous iterations and adjustments of the concept representation and therefore creates a significant bottleneck in the overall product development process.

The phase involves a meticulous process of iterating on the basic characteristics of the product design, considering various requirements including customer needs, technological capabilities, and physical constraints. The objective of the concept development is not merely to refine the ideas generated during the ideation phase but to explore and compare alternative solutions systematically. Each iteration serves as an opportunity to reassess and refine the product concept against the backdrop of evolving requirements and constraints. During the design conception phase, the visualization techniques play a crucial role in materializing abstract ideas into tangible representations. Whether through sketches, digital mock-ups, or physical prototypes, visualization aids in communicating ideas, facilitating discussions, and identifying potential design issues early in the development process. The exploration and comparison of solutions are instrumental in identifying the most viable path forward.

While block illustrations of functional structures, circuit diagrams and flow charts provide helpful illustrations, conceptual design relies on the two primary modalities text descriptions and visual representations. Visualizations of the product concept through sketches, images, and 3D models are vital as they relate to the products aesthetics and performance \cite{ulrichProductDesignDevelopment2008, benderPahlBeitzKonstruktionslehre2021}. In our work, we therefore focus on representations depicting the exterior design of product concepts.

\subsubsection{Product-Design Representations} \hfill

\textbf{2D-Shapes} play a vital role in the concept development of a variety of product designs as they are often used as initial, low-level parametric representations for finding feasible solutions. Through rapid ideation, they allow for experimentation with different shapes and are often part of the brainstorming process as they are easy to understand and manipulate. Furthermore, 2D-shapes are key in communication between designers, engineers, and stakeholders to share ideas and feedback. They additionally provide important insights about the feasibility of a geometry in early-stage simulations like aerodynamics, refer to \cite{chenAerodynamicDesignOptimization2019,chenInverseDesignTwoDimensional2022}. Despite their versatility, 2D-shape representations do not provide details about the semantics of the product design or about 3D characteristics, resulting in limited information about geometry and aesthetics.

\textbf{Images} serve as vital representations for product concepts in many domains. While they are used for structural and topology representations, they provide even more potential when used as natural image representations of the product design concepts, embodying intricate details and significantly contributing to the visual interpretation of product concepts. Key geometric and design aspects can be symbolized in a semantically meaningful manner. However, there are disadvantages to using images in design. For low-level design concepts, the pixel-based representation may lead to designs that are unsuitable or infeasible for further application. The consideration of 3D-features, both for semantic perception and performance assessment, demands intermediary transitions to 3D models since they are not provided in images. The disconnection between images and usable models, such as meshes or CAD models, remains a significant obstacle.

\textbf{3D-Objects} and geometries are often used as the go-to representation of product design concepts as they are the most information-rich approximation. Meshes are the most common method to represent objects in 3D space. They represent the geometry as a collection of connected vertices, edges and faces either as a surface or as a whole-body. They are utilized for downstream tasks like simulations but are challenging to synthesize, requiring specialized tools and software. Voxelizations depict the 3D equivalent to pixelizations in images and offer a volumetric representation. They can depict high-level and low-level features of the product design. For downstream tasks conversion into meshes is often required. Point-clouds are a collection of points in space representing the spatial distribution of the depicted object. Like Voxels, they often require conversion for downstream tasks.
     
\section{DGM-Driven Product Design Conception}
\label{sec:DGM_PDC}
\begin{figure*}
    \centering
    \includegraphics[width=0.8\textwidth]{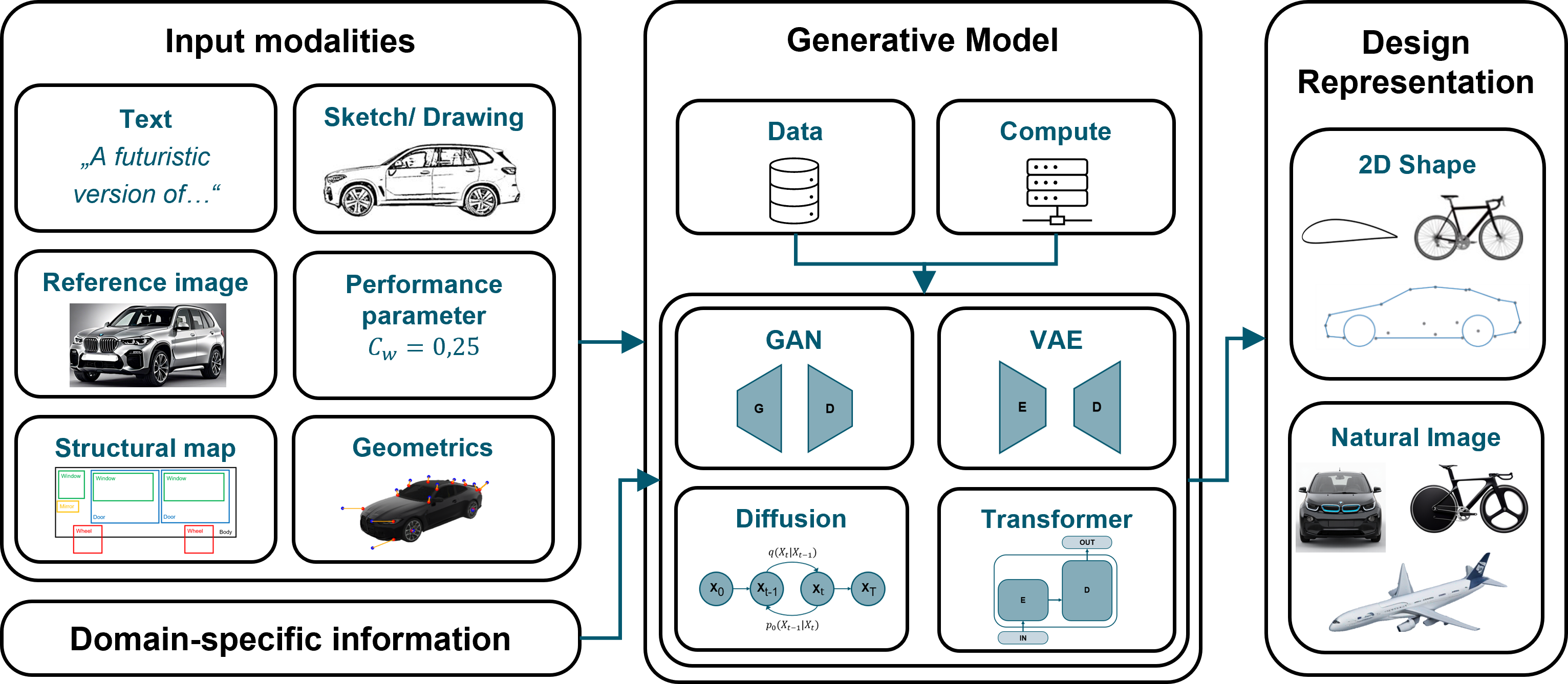}
    \caption{Essential components of DGM-driven generation process of Product Design Concepts.}
    \label{fig:DGM_driven_PDC}
\end{figure*}

Generative machine learning methods have made substantial progress in synthesizing complex data representations (text \cite{openaiChatGPT2023}, images \cite{rombachHighResolutionImageSynthesis2022, midjourneyinc.Midjourney2023},3D objects \cite{junShapEGeneratingConditional2023,shiZero123SingleImage2023}).They are met with increasing interest in the domain of engineering design \cite{alamAutomationAugmentationRedefining2024,joskowiczEngineersPerspectivesUse2023,heinCanMachinesDesign2018,picardConceptManufacturingEvaluating2023,regenwetterDeepGenerativeModels2022,zhangDeployingAINew2021}. 
The application of DGMs is especially promising in concept development \cite{zhangDeployingAINew2021} where text and visual representations are the primary modalities \cite{picardConceptManufacturingEvaluating2023}. We focus on PDC and therefore are especially interested in visual representations. A DGM-driven process to synthesize product design concepts is represented \Cref{fig:DGM_driven_PDC}.

\subsection{Potentials of DGMs in Product Design Conception}
The potential for AI-driven design is not just a theoretical possibility but a practical reality \cite{heinCanMachinesDesign2018}. DGMs are particularly interesting for their ability to rapidly generate a multitude of solutions for a single problem. This marks a paradigm shift from the traditional iterative approach that typically seeks a singular, optimal solution. This attribute of DGMs is especially beneficial in the realms of engineering- and PDC-processes, where the diversity of solutions can significantly enhance creativity and innovation \cite{regenwetterDeepGenerativeModels2022}.

The potential applications and benefits of DGMs in PDC are manifold. Key advantages include the reduction of costly late-stage design changes and providing critical information to designers and engineers by identifying suitable design spaces \cite{alamAutomationAugmentationRedefining2024}. Traditionally, iterative design workflows heavily rely on the domain expertise of skilled engineers, necessitating a substantial investment of time and effort. These workflows are often fractured across multiple specialized teams, leading to inefficiencies through iterative hand-offs. Constraints need to be communicated back and forth among various experts. In contrast, DGMs offer the promise of streamlining this process by producing designs that satisfy constraints earlier in the design lifecycle, thereby facilitating a more cohesive and efficient approach to product development.

Moreover, DGMs allow for the rapid visualization of concepts and ideas, which can lead to better representations of customer needs and an improved understanding of how basic performance requirements impact the product design. This shift towards a constraint-driven design process enables designers to focus more on product performance and constraints \cite{saadiGenerativeDesignReframing2023}, fostering a more exploratory approach to considering a wider range of possible product solutions \cite{songWhenFacedIncreasing2022}.

\subsection{Open Challenges for DGM-Application in PDC}
While DGMs herald a new era of potential in product design generation, their successful integration into user-centric processes reveals fundamental challenges. Despite the evident potential of DGMs to revolutionize conceptual design, their widespread adoption has been relatively slow for various reasons. While there is a general willingness to incorporate and adapt these methods, particularly for text-based applications, hesitations persist. Concerns stem from uncertainty about the outcomes, a lack of knowledge among potential users, and questions regarding the precision of the generated representations \cite{joskowiczEngineersPerspectivesUse2023}.
The gap between the theoretical promise of DGMs and their practical application in PDC is facilitated by the struggle of current DGMs with tasks requiring advanced spatial reasoning and the solving of complex design problems. This limitation not only undermines the consistency and feasibility of the generated outputs but also diminishes their utility in practical design settings \cite{picardConceptManufacturingEvaluating2023}. The prevailing shortcomings of DGMs in this context include their inability to generate robust, reliable, and replicable outputs. This is compounded by a lack of relevant domain knowledge, an unawareness of industry standards, difficulties in integrating with existing workflows, and challenges in interpreting data from diverse sources and formats \cite{alamAutomationAugmentationRedefining2024}. Further compounding these issues are the challenges associated with ensuring that generated designs adhere to explicit design constraints, consider design performance, ensure physical feasibility, promote design novelty, and navigate data sparsity \cite{regenwetterDeepGenerativeModels2022}.

\subsection{Requirements for DGM-Application in PDC}
Promises of DGMs in PDC are underwhelming unless the limitations, outlined in the previous sections, are addressed. Addressing the existing gaps presupposes understanding and acknowledgment of the inherent requirements and obstacles that currently curtail the full exploitation of DGM-capabilities in this domain. Key requirements for DGMs in engineering design include reliability, with models delivering consistent performance; stability, ensuring predictable functioning under a variety of conditions; accuracy, providing outputs that are both precise and controllable; adaptability, allowing for seamless adjustment to new or evolving requirements; and a strict adherence to predefined product specifications and design constraints \cite{alamAutomationAugmentationRedefining2024}. 
The incorporation of DGMs into complex design problems yields tangible benefits only when the context and interactions between human designers and AI-models are made intuitively accessible \cite{zhangCautionaryTaleImpact2021}. \cite{khanHowDoesAgency2023} advocate for a collaborative approach wherein both humans and autonomous agents explore a diverse set of designs that align with human preferences. This approach seeks to strike a balance between the performance and novelty of solutions, suggesting that the synergy between human intuition and machine efficiency is crucial for early design exploration.

Drawing from the existing body of research on the capabilities and shortcomings of DGMs and informed by profound theoretical models on the product development process \cite{ulrichProductDesignDevelopment2008,benderPahlBeitzKonstruktionslehre2021}, we propose a set of core requirements for the effective incorporation of DGMs into the development of product concepts, summarized in \Cref{tab:Requirements}. These requirements aim to address the highlighted challenges by fostering a more seamless integration of DGMs into the design process, ensuring that the generated outputs are not only innovative and diverse but also practical, feasible, and aligned with the intricate web of constraints and standards that define the field of product design.

\subsubsection{Conditioning and Controllability}
To be applied in the generation of product design concepts, DGMs must allow for conditional controllability of the generative process by the user. User input for generative tasks has to be intuitive, easy to edit interactively, and commonly used in the traditional creative process providing modalities to incorporate specifications about the products purpose, distinct visual, functional, and technical features, and performance requirements \cite{xueDeepImageSynthesis2022}. Relevant modalities to formulate the objective of the generative task come in the form of text, visual inputs, and performance parameters.

\textbf{Natural Language Text} is the most intuitive way of human communication and allows users to articulate specific requirements and demands in simple or more complex formulations. It allows for flexible articulation of complex ideas, requirements, and preferences. While natural language can capture high-level concepts and abstract ideas (“futuristic”, “inspired by nature”) to assist the synthesis of novel and unconventional content, it requires sophisticated models for processing and is inherently ambiguous, as the same text-prompt can have a different meaning depending on the observer.

For visual conditionings, \textbf{Reference Images} provide the possibility of using existing image representations as guidance for the generative process. They allow for the utilization of existing designs and concepts as visual references for type, geometry, style or features of the product design concept and therefore create a starting point for creativity and design exploration.  

\textbf{Sketches and Drawings}, depicting contours and cross-sections of the product design, are a crucial part of early-phase design exploration and therefore relevant modality for conditioning. They allow to represent fundamental ideas and features in a low-fidelity format that contains relevant visual or geometric details and is easy to understand and manipulate. \textbf{Structural Inputs} (Edge-, Depth-, Spatial- and Segmentation-Maps) depict spatial and geometrical characteristics like edges, segmented objects, depth fields and bounding boxes to describe the spatial composition of the target representation. They allow for explicit control over the spatial composition and characteristics of the design concept, facilitating the inclusion of spatial information product features.

\textbf{Performance parameters} ensure that specific performance criteria are met and help to ensure technical feasibility of the design. They come in the form of numerical values or dimensionless ratios (e.g.: lift coefficient of airfoils, drag coefficient of vehicle silhouettes). Performance parameters may limit the creative exploration if too narrowly defined.

\subsubsection{Consistency}
\begin{figure*}
\centering
\begin{subfigure}{0.49\textwidth}
    \centering
    \includegraphics[width=\linewidth]{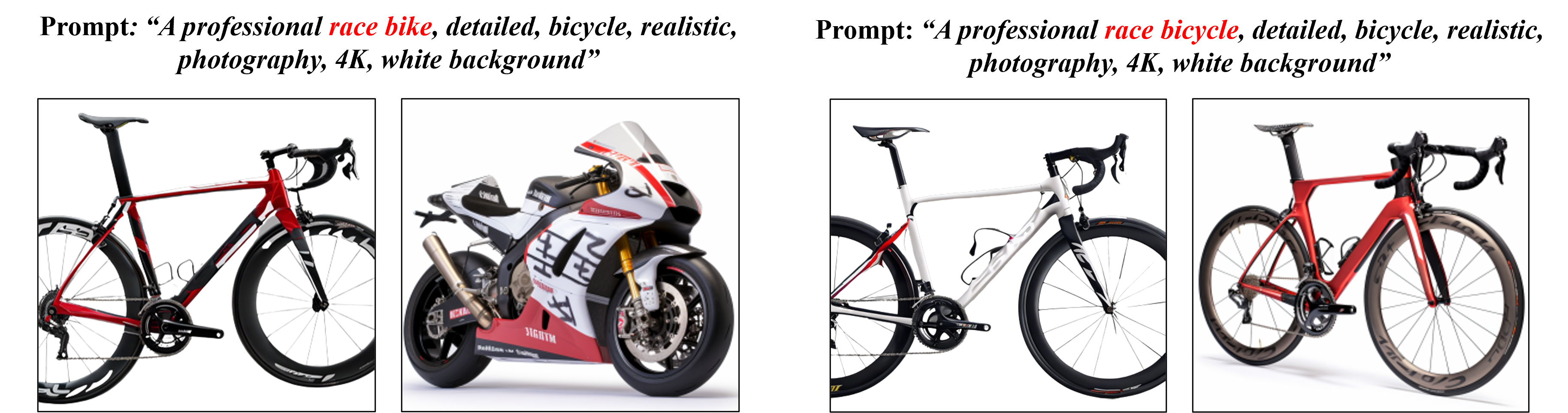}
    \caption{Inconsistency of the generated image (left: SD \cite{rombachHighResolutionImageSynthesis2022}; right: Midjourney \cite{midjourneyinc.Midjourney2023}.)}
    \label{fig:Inconsistencies} 
\end{subfigure}
\hfill
\begin{subfigure}{0.49\textwidth}
    \centering
    \includegraphics[width=\linewidth]{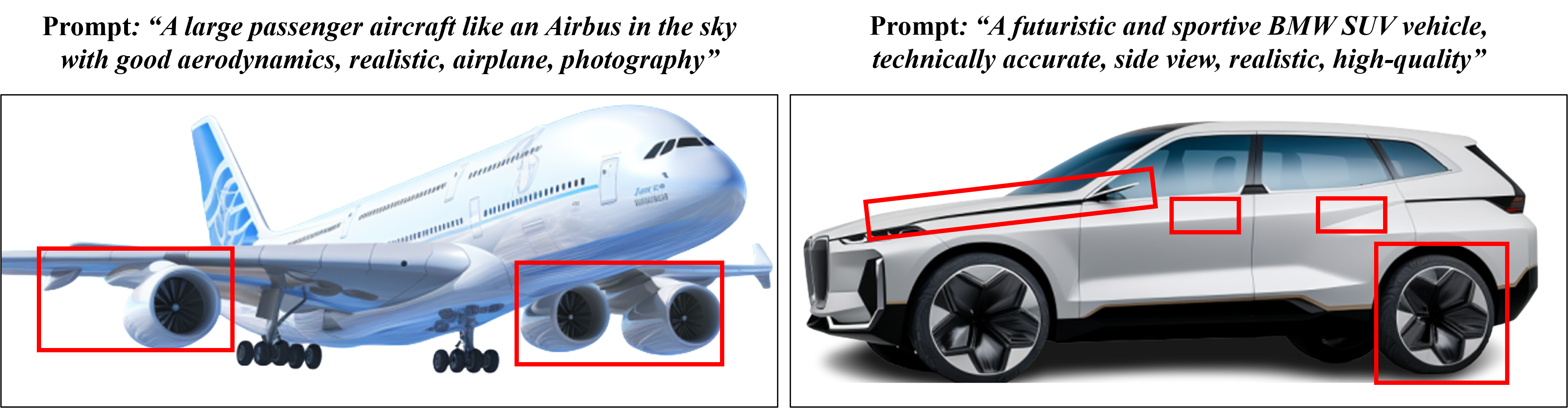}
    \caption{Failure cases of real-world coherence (generated with Midjourney \cite{midjourneyinc.Midjourney2023}.}
    \label{fig:Incoherence}     
\end{subfigure}
\caption{Examples of visual DGMs not fulfilling the requirements for PDC.}
\label{fig:Failure_examples}
\end{figure*}

Inconsistencies between the generated output and the users intentions remain an open challenge in DGMs and headwind their application in product and engineering design \cite{alamAutomationAugmentationRedefining2024}. They originate from various sources; a prominent example are ambiguities in the conditioning (\Cref{fig:Inconsistencies}). Ensuring consistency in the output generated by DGMs in relation to user inputs and conditions is a critical requirement for their application in PDC.

For the generation process to be considered truly controllable, it is imperative that the design concepts depict the main geometric features and design characteristics essential for the concept at hand. User-provided inputs and conditionings, whether they are specific design parameters, aesthetic preferences, or functional criteria, need to be respected by the DGM. Ultimately, the sophistication of the model and the conditioning mechanisms employed are rendered moot if the outputs fail to align with the user's intentions.

This alignment is what enables DGMs to serve as effective and efficient tools in the hands of designers, facilitating the creative process by ensuring that the conceptual outputs are consistently in tune with the initial design

\subsubsection{Coherence}
The probabilistic nature of most current-generation DGMs often contradicts the general requirement of real-world coherence, as the models have learned the probability distribution of the data, not physical laws, as depicted in \Cref{fig:Incoherence}.

Coherence ensures that the generated outputs not only adhere to basic physical principles, showcasing an elementary level of world knowledge, but also align with domain-specific functional requirements. This dual alignment imbues the generated designs with a sense of realism and practical applicability, making them more than mere imaginative explorations, which is essential to be considered seriously within the engineering process and to contribute effectively to the innovation pipeline. To bridge the creativity gap by offering variations and novel designs that are desirable and valuable, it is not enough for generated solutions to merely replicate training data. DGMs need to produce outputs that are both innovative and aligned with real-world constraints and possibilities.

\subsubsection{Customization}
For DGMs to be truly effective, they must not only generate outputs that are coherent with general physical and technical principles but also embody the nuanced attributes unique to their specific domain. This requirement for customization necessitates the efficient induction of further domain-specific data and knowledge into the models, either through fine-tuning mechanisms or as conditional inputs. This process allows for the better utilization of existing information, ensuring that the generated designs are not just plausible but also aligned with domain-specific requirements. This requirement can be viewed as an additional layer of detail to the need for coherence as it aims to ensure that the models produce results that are both universally plausible and tailored to the unique demands of the domain.

In PDC, the ability to use domain-specific inputs and understand their implications is critical as specifications are often subject to specific sets of conventions \cite{alamAutomationAugmentationRedefining2024,picardConceptManufacturingEvaluating2023}. Generative models must be adaptable enough to interpret these inputs correctly and generate outputs that conform to these specialized communication forms. 

\subsubsection{Data and Computational Costs}
The proprietary nature of most engineering design data, coupled with strict intellectual property rules [13], poses a significant challenge in aggregating the amounts of data typically required for training DGMs. The question of computational cost and data efficiency is not just a technical hurdle but a critical factor in the broader applicability of DGMs in domain-specific PDC. In many instances, the feasibility of training a DGM on millions of data points, utilizing hundreds of hours on high-end GPUs, is simply not practical within the constraints of industry-specific applications. Domain-specific applications often grapple with limitations not just in terms of the volume of training data—frequently only a few thousand data points—but also in the available computational resources, which may extend only to consumer-grade GPUs. This scenario underscores a critical challenge: achieving computational efficiency and managing the substantial costs associated with deploying DGMs that demand extensive data and compute for training.

Addressing these challenges necessitates solutions, such as the potential to fine-tune pre-trained models or the development of small-scale adapters capable of learning from modest amounts of highly specific data. Such approaches enable the customization of large, general-purpose models to generate high-quality, domain-specific representations with a fraction of the data and computational power typically required. 

\begin{table*}
\small
\caption{Requirements for DGM-application in Product Design Conception.}
    \centering
    \begin{tabularx}{\textwidth}{ZY}
        \toprule
        \textbf{Requirement} & \textbf{Description} \\
        \midrule
        Controllability & \textit{Ability to control the generated content with user-provided conditioning inputs of relevant modalities. E.g.: Text-prompt, Sketch, Performance Parameter, Reference Image}\\
        \midrule
        Consistency	& \textit{Consistency of the generated content with the provided input conditions and constraints.}\\
        \midrule
        Coherence & \textit{Coherence of the generated content with real-world characteristics of the product and general physical and technical laws.}\\
        \midrule
        Customization & \textit{Ability to incorporate domain-specific information or characteristics into the generative process to customize the outputs.}\\
        \midrule
        Cost & \textit{Availability of required amounts of data and justifiability of required amounts of computational resources.}\\
        \bottomrule
    \end{tabularx}
\label{tab:Requirements}
\end{table*}

\section{Analysis of DGM-Applicability in Product Design Conception}
\label{sec:Analysis}

In this chapter, we conduct a technical analysis of DGMs to assess their applicability and potential utility in PDC. We direct readers more interested in the implications of this analysis on the applicability in PDC to \Cref{sec:recommend}.

Focusing on Variational Autoencoders (VAEs), Generative Adversarial Networks (GANs), Diffusion Models, Transformers, and Radiance Field Methods, we scrutinize these models through the lens of the five requirements identified in the preceding chapter: Controllability, Consistency, Coherence, Customization, and Cost (Data and Compute). Our objective is twofold: firstly, to delineate the strengths and weaknesses of each model family in meeting these requirements, thereby guiding engineers and designers in selecting the most suitable model family for their specific design tasks; secondly, to identify areas where further research could enhance the applicability and effectiveness of DGMs in engineering and PDC.

\subsection{VAE's}
VAEs are characterized by their capability of embedding input data into an interpretable, lower-dimensional latent space \cite{kingmaAutoEncodingVariationalBayes2013}, facilitating smooth interpolation between points in the learned distribution. Their architecture balances reconstruction accuracy with adherence to the latent space distribution \cite{kingmaIntroductionVariationalAutoencoders2019}. This capability is particularly beneficial in domains requiring a grasp of the underlying data distribution. They are generally less prone to training instabilities, making them user-friendly for non-experts. Relevant representations where VAE’s have been applied are 2D-Shapes \cite{yonekuraGeneratingVariousAirfoil2021, burnapEstimatingExploringProduct2016} and 3D-Objects \cite{lennonImage2LegoCustomizedLEGO2021,nozawa3DCarShape2022,sanghiCLIPSculptorZeroShotGeneration2023,lipredictivegenerativedesign2022,fuShapeCrafterRecursiveTextConditioned2023}.

\textit{1. Controllability:}
By nature, VAEs operate by sampling randomly from a learned probability distribution, an approach that is inherently unconditional and thus does not provide direct control over the specifics of the generated content. To imbue VAEs with a degree of controllability, conditional sampling must be explicitly incorporated during the training phase \cite{sohnLearningStructuredOutput2015}. Achieving highly detailed control over specific features or high-resolution outputs often falls beyond their current capabilities.
Possible conditioning modalities span a range of inputs, including text, which can be effectively utilized through the integration of models like CLIP \cite{radfordLearningTransferableVisual2021} as demonstrated in \cite{fuShapeCrafterRecursiveTextConditioned2023,sanghiCLIPSculptorZeroShotGeneration2023}; visual inputs, such as sketches and images \cite{lennonImage2LegoCustomizedLEGO2021,lipredictivegenerativedesign2022,nozawa3DCarShape2022} and performance parameters \cite{yonekuraGeneratingVariousAirfoil2021}.

\textit{2. Consistency:}
The requirement for consistency between user inputs and the generated outputs in VAEs presents a complex challenge, due to the intrinsic probabilistic nature of these models. This inherent characteristic often stands in contrast to the precise expectations set by conditioning inputs. Incorporating conditioning into the learned probability distribution aims to guide the generative process towards outputs that align more closely with specified parameters. However, the decoding process, which translates latent representations back into comprehensible outputs, is notoriously difficult to control. This difficulty underscores the challenge in enforcing strict constraints within the generative process, often resulting in outputs that diverge from the intended specifications.

\textit{3. Coherence:}
The continuous representation of data in the latent space, while facilitating smooth interpolations, may inadvertently encompass areas that yield designs and interpolations deemed physically infeasible or technically implausible. Classical-architecture VAEs have tendency to produce blurry outputs in tasks demanding high-resolution detail \cite{burnapEstimatingExploringProduct2016}. Such blurrier results directly impact the model's ability to accurately reflect detailed physical and technical nuances of designs. Some VAE-based approaches aim to generate high-quality variations of product and object designs, but are limited due to the lack of explicit user-conditioning capabilities \cite{zhang3DShapeSynthesis2019,parkDeepSDFLearningContinuous2019}. 

To achieve higher quality and more coherent results, adaptations to the model's architecture, such as incorporating adversarial training elements, have been explored with promising outcomes \cite{fanAdversarialLatentAutoencoder2023}.
To increase the generalization and coherence of VAE outputs with real-world conditions, novel approaches continue to be investigated \cite{geMetaConditionalVariational2022}. Attempts to address these issues have seen success through the adoption of discrete representations of the data in the latent space \cite{oordNeuralDiscreteRepresentation2018}. This approach not only scales better but also enhance the model's capacity to manage complex, high-level data such as images, thereby improving coherence with real-world expectations.

\textit{4. Customization:}
VAEs' ability to handle medium sizes of data relatively well makes them generally feasible for training from scratch on domain-specific datasets. This feasibility translates to less computational expense compared to training more data-intensive models, offering a practical advantage for customization to specific engineering or design tasks. However, this customization comes with a notable trade-off: a model highly customized for specific domain tasks often experiences a loss in its generalization capabilities. This trade-off is contingent upon several factors, including the complexity of the model itself and the size and format of the data used for training. Tailoring a VAE too closely to a particular set of data or task requirements may limit its applicability across broader or slightly different tasks within the same domain.

\textit{5. Cost (Data and Compute):}
VAEs are known for their relatively modest computational requirements compared to other DGMs This efficiency makes VAEs particularly attractive for scenarios with limited computational budgets or where rapid prototyping is necessary. The feasibility of training VAEs on consumer-level hardware underscores their practicality. \cite{zhang3DShapeSynthesis2019} demonstrated that training a VAE-model to generate domain-specific 3D-Objects could be completed within 3 hours using a Geforce 1080 GPU. 
However, it's essential to balance the expectations regarding the quality of the output and the complexity of the task with the computational resources available. While VAEs can be trained on consumer-grade hardware, the resolution and complexity of the generated designs might be constrained by the computational power and the size and quality of the dataset used for training. 

\subsection{GAN's}
Generative Adversarial Network have established themselves as powerful tools to generate complex data structures like images. Their proficiency in producing realistic images surpasses that of VAEs. GANs leverage an adversarial training process, characterized by the dynamic competition between the generator and discriminator \cite{goodfellowGenerativeAdversarialNets} that significantly enhances the fidelity of generated designs \cite{karrasStyleBasedGeneratorArchitecture2021}. This process fosters a continual improvement in the quality of generated outputs and has been extended to general-purpose image generation tasks \cite{karrasAnalyzingImprovingImage2020,sauerStyleGANTUnlockingPower2023,sanghiCLIPSculptorZeroShotGeneration2023}. GANs have also found various applications in product design. They are used to generate 2D-aerodynamic shapes \cite{chenInverseDesignTwoDimensional2022,nobariPcDGANContinuousConditional2021,chenBezierGANAutomaticGeneration2021} and structural designs of vehicle wheels \cite{ohDesignAutomationIntegrating2018}.
Due to the adversarial training process, GANs are notoriously difficult to train. Issues such as mode collapse (the model overfits to a subspace of the solution space) and the potential for either the generator or discriminator to become disproportionately powerful, lead to instabilities and high sensitivity on training conditions.

\textit{1. Controllability:} GANs have been adapted to incorporate various conditioning modalities, since a general conditioning methodology was introduced \cite{mirzaConditionalGenerativeAdversarial2014}. The modalities range from performance parameters like Mach number and lift coefficient \cite{chenInverseDesignTwoDimensional2022} to reference images  \cite{ohDesignAutomationIntegrating2018} and textual descriptions \cite{sauerStyleGANTUnlockingPower2023,maGenerativeAdversarialNetwork2023,couaironFlexITFlexibleSemantic2022}. Image editing tasks have also been covered, allowing for inpainting \cite{isolaImagetoImageTranslationConditional2017}, image synthesis from segmentation and edge maps \cite{deng3DawareConditionalImage2023}, and point-dragging mechanisms \cite{panDragYourGAN2023}. 
Notably, the conditioning in GANs must be incorporated from the onset of the training process, which inevitably introduces increased complexity and potentially higher computational costs. 

\textit{2. Consistency:} The adversarial training process inherent to GANs plays a pivotal role in ensuring consistency between user-provided inputs and the generated content, a feature that often positions GANs favorably in comparison to VAEs. This inherently incentivizes the production of outputs that closely match the conditioning inputs. Mismatches between conditioning inputs and generated images can be further addressed by incorporating a semantic comparison module \cite{maGenerativeAdversarialNetwork2023}, which maps both the input texts and the generated images into a shared semantic latent space, allowing for a direct comparison to ensure alignment. The introduction of a hybrid attention mechanism in the generator serves to direct the focus of the model on relevant aspects of the input, thereby enhancing the fidelity of the generated content to the specified conditions.

\textit{3. Coherence:} At their core, GANs are incentivized to remain close to the distribution of their training data. While GANs can produce outputs that exhibit a high degree of coherence within the realms they have been trained on, their ability to generalize and maintain this coherence in out-of-distribution scenarios is inherently constrained. The generation of content that accurately reflects real-world knowledge and feasibility necessitates vast amounts of training data, covering a broad spectrum of scenarios and contexts. The introduction of dynamic Gaussian mixture latents into the GAN generator is designed to increa\cite{yangDFSGANIntroducingEditable2023}. Vector quantization has also shown to increase quality and realism in GANs \cite{esserTamingTransformersHighResolution2021}, especially in combination with text encoding \cite{changMuseTextToImageGeneration2023}.

\textit{Customization:} The necessity to tailor GAN models to particular domains can be addressed with several approaches. Few-shot generation utilizes a small set of reference images as conditioning, enabling the model to generate new images that adhere to the domain-specific characteristics reflected in the limited dataset \cite{robbFewShotAdaptationGenerative2020}. Fine-tuning pre-trained GAN models offers another avenue for customization \cite{moFreezeDiscriminatorSimple2020}, adapting the model to new domains using relatively small amounts of data. In general, the topic of directing the generative capabilities of GANs towards specific domains using limited amounts of data has already been covered by numerous studies. We therefore direct the interested reader to the comprehensive overview by \cite{yangImageSynthesisLimited2023}.

\textit{5. Cost (Data and Compute)} In domain-specific applications, GANs have shown efficiency. \cite{chenBezierGANAutomaticGeneration2021} achieved training times between 6 and 13 minutes on an NVIDIA Titan X, a consumer-level GPU, with data requirements ranging from 5000 to 10000 datapoints. Similarly, \cite{ohDesignAutomationIntegrating2018} managed to train their model with just 1728 datapoints for a few hours on an Nvidia GTX 1080, and \cite{chenInverseDesignTwoDimensional2022} completed their training between 2 and 4 hours on a Nvidia Tesla V100. Contrastingly, the training of general-purpose models demands significantly higher computational resources and data. StyleGAN-T reported training durations of four weeks on 64 Nvidia A100 GPUs—although notably still only a quarter of the resources used by models like Stable Diffusion—and required an extensive dataset of 250 million datapoints \cite{sauerStyleGANTUnlockingPower2023}. 
Training instabilities remain a hurdle across the board, impacting both the efficiency and effectiveness of GANs in practice. We direct those interested in delving deeper into the nuances of GAN training efficiency to the study of \cite{yangImageSynthesisLimited2023}.

\subsection{Diffusion Models}
Diffusion models have rapidly ascended to the forefront of DGM-based image synthesis, achieving unparalleled success in producing high-quality results that outperform GANs across numerous benchmarks \cite{dhariwalDiffusionModelsBeat2021}. State-of-the-art models include Midjourney \cite{midjourneyinc.Midjourney2023}, Stable Diffusion \cite{rombachHighResolutionImageSynthesis2022}, and DALL-E 3 \cite{betkerImprovingImageGeneration2023}. Diffusion models have also been applied for generative tasks involving 3D-Point-Clouds \cite{nicholPointESystemGenerating2022,zengLIONLatentPoint2022,zhou3DShapeGeneration2021} iteratively refining the generated objects in multiple steps. Additionally, there has been recent progress in generating 3D-Objects with diffusion-based models \cite{shiZero123SingleImage2023,liuZero1to3ZeroshotOne2023,voletiSV3DNovelMultiview2024}. A comprehensive overview on the current state-of-the-art of diffusion models, their capabilities in image-, video- and 3D-generation, conditioning mechanisms and customization is given by \cite{poStateArtDiffusion2023}, which we strongly recommend for the interested reader.

Despite their impressive generative prowess, diffusion models come with significant drawbacks, primarily their substantial demands for data and computational resources for training and inference. This requirement poses challenges, especially for researchers and practitioners with constrained access to resources. Moreover, while there is a concerted effort within the community to enhance computational efficiency, the technical complexity involved in modifying and building adapters for these models necessitates a deep understanding of their architecture. Fortunately, the community is characterized by its widespread open-source ethos and strong effort to democratize access through user-friendly implementations. Numerous powerful, general-purpose models are available for fine-tuning and adaptation in libraries like Diffusers on HuggingFace \cite{vonplatenDiffusersStateoftheartDiffusion2024}. Continuous developments and research are aiming to address the current limitations of diffusion models, shown by the steep increase in research activities in the field (see Figure 7). 

\textit{1. Controllability:} The standard implementation of diffusion models operates in an unconditional manner, generating outputs based solely on learned distributions. However, a variety of methodologies for introducing controllability have been developed recently. Text conditioning has become prevalent, due to its intuitive usability and the advent of pretrained text-image encoders like CLIP \cite{radfordLearningTransferableVisual2021}. Notable examples of large-scale text-to-image models include Stable Diffusion \cite{rombachHighResolutionImageSynthesis2022} and its successor SDXL \cite{podellSDXLImprovingLatent2023}; Imagen \cite{sahariaPhotorealisticTexttoImageDiffusion2022}, Dall-E \cite{betkerImprovingImageGeneration2023} and Midjourney \cite{midjourneyinc.Midjourney2023}. Visual inputs have also been widely researched. Modalities range from reference images \cite{changMuseTextToImageGeneration2023,nicholGLIDEPhotorealisticImage,galImageWorthOne2022,ruizDreamBoothFineTuning2023}, to sketches and drawings\cite{zhangAddingConditionalControl2023,voynovSketchGuidedTexttoImageDiffusion2022}, structural inputs \cite{zhaoUniControlNetAllinOneControl2023,bar-talMultiDiffusionFusingDiffusion2023} and geometric inputs \cite{shiDragDiffusionHarnessingDiffusion2023,mouDragonDiffusionEnablingDragstyle2023}.

For the injection of conditioning information into the generative process, attention mechanisms play a key role. Stable Diffusion, ControlNet and Zero-1-to-3 utilize cross-attention mechanisms, while Zero-123++ uses reference attention \cite{rombachHighResolutionImageSynthesis2022,shiZero123SingleImage2023,liuZero1to3ZeroshotOne2023,zhangAddingConditionalControl2023}. In general, the technical complexity as well as the computational cost for conditioning varies greatly between approaches, requiring careful analysis of the specific requirements.

\textit{2. Consistency:} Despite numerous conditioning mechanisms and efforts to improve their accuracy, ensuring consistency between user inputs and the generated content remains a challenge. The inherent nature of the diffusion process poses obstacles to achieving precise consistency due to iterative, probabilistic sampling. Controlling the latent representations of the generated contents and the denoising process is possible but challenging. Ambiguities in the conditioning input are another source of error (\Cref{fig:Failure_examples}). Eliminating these leads to increased consistency of the generated content \cite{betkerImprovingImageGeneration2023,yangMasteringTexttoImageDiffusion2024}. Other approaches include iterative alignment of the representation with the user-defined target \cite{shiDragDiffusionHarnessingDiffusion2023} and analysis or modification of the initial noise, from which the denoising process starts \cite{mouDragonDiffusionEnablingDragstyle2023,bansalColdDiffusionInverting2022}.

\textit{3. Coherence:} Diffusion models do not intrinsically possess an understanding of physical laws and technical constraints, as seen in Figure 4. The lack of contextual reasoning and relating between parts and objects in the representations are well-known challenges in diffusion models that require explicit attention. Coherence and a sense of world knowledge are gradually acquired as large-scale models are exposed to extensive examples across diverse concepts \cite{openaiVideoGenerationModels2024}.

There are several approaches to address the specific technical constraints and ensure feasible outputs, one being the concatenation of additional channels that containing physical properties to the image data when training the diffusion model \cite{shuPhysicsinformedDiffusionModel2023}. 

Other approaches aim to modify and optimize the noise schedule and the sampling to enforce feasible results \cite{shiZero123SingleImage2023,maoGuidedImageSynthesis2023,fanNoiseSchedulingGenerating2023}. Finetuning and preference optimization methods can also yield increased coherence in domain-relevant areas by providing the model with more context \cite{clarkDirectlyFineTuningDiffusion2023,wallaceDiffusionModelAlignment2023}.

Finally, interpreting and modifying the latent dimensions of diffusion models, an approach originally proposed to enhance user safety \cite{liSelfDiscoveringInterpretableDiffusion2023,haasDiscoveringInterpretableDirections2023}, offers another pathway to constrain and guide the generation process towards more coherent outputs.

\textit{4. Customization}: Given the prohibitive computational and data requirements of training large-scale, general-purpose diffusion models from scratch, more efficient strategies for domain-adaptation are highly sought after. Injecting domain-specific information into pre-trained models, thereby enabling customization, can be done through finetuning the models’ weights. Low-Rank Adaptation, a technique initially proposed for large language models (LLMs) \cite{huLoRALowRankAdaptation2021,dettmersQLoRAEfficientFinetuning2023}, has been adapted for use in diffusion models and provides a more efficient alternative to classical finetuning that updates all weights. Other finetuning me\cite{clarkDirectlyFineTuningDiffusion2023,qiuControllingTexttoImageDiffusion2023}.

Few-shot learning techniques to introduce specific concepts into the model, lowering the data threshold for effective customization and providing an option for tailoring the model's output with just a handful of examples \cite{galImageWorthOne2022,ruizDreamBoothFineTuning2023}. Relying on only a single example, One-Shot Identity Preservation, is currently implemented primarily for human faces \cite{wangInstantIDZeroshotIdentityPreserving2024}. Another strategy involves adjusting the implicit assumptions within diffusion models. Updating the ingrained biases may help to make models more accurate and direct them to the specific standards and nuances of the target domain \cite{orgadEditingImplicitAssumptions2023}. Although promising in theory, effectiveness and practical applicability of these approaches need to be investigated in more detail.

\textit{5. Cost (Data and Compute):} The computational cost and data requirements of diffusion models pose significant challenges to their widespread application, particularly in scenarios where resources are limited. The training of Stable Diffusion on 256x256 ImageNet data exemplifies this challenge, having consumed approximately 271 V100 GPU days \cite{rombachHighResolutionImageSynthesis2022}. Other models of that scale can be approximated to have consumed similar resources. For generating 3D-objects, the computational costs are at least as high and likely exceed the cost for image-models. Stability AI’s SV3D model \cite{voletiSV3DNovelMultiview2024}, that utilizes latent video diffusion, trains for a sum of 1000 A100 GPU days, when trained from scratch.

The investments needed for conditioning mechanisms vary significantly based on the desired level of generalization and the implementation approach. Training a ControlNet adapter demands several hundred hours on an A100 GPU \cite{zhangAddingConditionalControl2023}, while other methods present a more resource-efficient alternative, requiring merely about an hour on a single A100 GPU \cite{voynovSketchGuidedTexttoImageDiffusion2022}. Some conditioning and modification mechanisms circumvent the need for additional training altogether but necessitate a deep understanding of the model's inner workings and careful experimentation to achieve desired outcomes effectively \cite{shiDragDiffusionHarnessingDiffusion2023}. The choice among these options requires careful consideration of the available resources, the specific goals of the project, and the level of technical expertise.

\subsection{Transformers}
Transformers are the dominant architecture in natural language processing with unparalleled capabilities in language understanding and generation \cite{vaswaniAttentionAllYou2017}. Their success in sequence modeling has paved the way for applications involving visual tasks \cite{dosovitskiyImageWorth16x162021,childGeneratingLongSequences2019}. By treating the data as sequences, transformers offer a high level of explicit control over the generation process that is particularly advantageous for applications demanding high precision. They have been applied for tasks involving image generation \cite{esserTamingTransformersHighResolution2021} and 3D-Object synthesis \cite{siddiquiMeshGPTGeneratingTriangle2023}. Recent developments like the Diffusion Transformer, a hybrid architecture improving the methodology of diffusion models with transformers, and rectified flow transformers increase the generative capabilities in visual tasks \cite{peeblesScalableDiffusionModels2023,crowsonScalableHighResolutionPixelSpace2024,esserScalingRectifiedFlow2024}. Transformer architectures have also been applied to specifically model CAD-based mechanical design processes \cite{wuDeepCADDeepGenerative2021,xuBrepGenBrepGenerative2024,xuSkexGenAutoregressiveGeneration2022,jayaramanSolidGenAutoregressiveModel2023}.

Transformers cover a much broader spectrum of applications. Atop the resource-intensive nature of training that requires substantial datasets and computational power, selecting an efficient and feasible Transformer-based architecture poses a significant challenge.

\textit{1. Controllability:} Predominantly, transformer-based models have embraced text-conditioning as the primary method for conditioning \cite{junShapEGeneratingConditional2023,sanghiCLIPSculptorZeroShotGeneration2023,rameshZeroShotTexttoImageGeneration2021}. Classic masked-inpainting, as well as mask-free, text-guided inpainting have also been explored \cite{changMuseTextToImageGeneration2023}. Beyond text, the spectrum of controllability includes depth maps, structural maps, and class conditions \cite{esserTamingTransformersHighResolution2021}. Pioneering an approach by conditioning generation on technical sketches traditionally used in CAD applications, efforts have been made to bridge the gap between abstract design concepts and practical engineering applications \cite{xuSkexGenAutoregressiveGeneration2022}. Compared to diffusion models, there exist substantially fewer studies on conditioning transformer-based visual models. The requirement for large amounts of computational resources and advanced technical knowledge to effectively train, adapt and condition these models poses barriers to accessibility and practical implementation.

\textit{2. Consistency:} Transformer-based visual models are just beginning to carve out a significant presence, with substantial advancements being made only recently. This stage inherently comes with uncertainties, particularly in ensuring that outputs consistently match the specificity and complexity of user inputs, a challenge that is magnified in domain-specific applications. Preliminary studies find that transformer-based visual models can achieve higher levels of cardinality, better adherence to the compositionality of conditioning inputs, and improved text-rendering capabilities compared to their state-of-the-art diffusion model counterparts \cite{changMuseTextToImageGeneration2023}. This suggests a potential for these models to offer more reliable and consistent results, especially as they scale \cite{yuScalingAutoregressiveModels2022}. As the size of the models expands, so too does their ability to adhere to user inputs in the generated outputs.

While this trend offers a glimpse into the potential of transformer-based, the practicality of scaling these models presents its own set of challenges, particularly in terms of computational demands and technical expertise required.

\textit{3. Coherence:} Transformer-based visual models exhibit promising potential for enhancing real-world coherence in generated content. Unlike diffusion-based models, where coherence is mainly influenced by the conditioning, they inherit increased contextual capabilities and the ability to relate between objects in an image, improving overall coherence \cite{gaoMDTv2MaskedDiffusion2024}. There is an aspiration to mirror the scalability and zero-shot performance observed in Large-Language Models, by focusing on predicting the next resolution scale rather than the next token in a tokenized image. Preliminary results reveal promising strides toward achieving more coherent and scalable image generation capabilities \cite{tianVisualAutoregressiveModeling2024}. Enabling bidirectional information flow between image and text tokens significantly enhances text comprehension and the semantic alignment of generated images with human preferences \cite{esserScalingRectifiedFlow2024}.

Real-world coherence achieved by transformer-based models is contingent on further scaling and development, evident in the Diffusion-Transformer architecture that demonstrates promising adherence to physical laws when scaled \cite{openaiVideoGenerationModels2024}. While early results are promising, comprehensive research and validation are necessary to balance cost and efficiency, especially for domain-specific applications. 

\textit{4. Customization:} The customization of transformer-based visual models for domain-specific applications represents a relatively underexplored area, particularly due to the novelty and computational intensity associated with these architectures. Unlike their textual counterparts, which have seen extensive customization through fine-tuning across a multitude of tasks \cite{huLoRALowRankAdaptation2021,dettmersQLoRAEfficientFinetuning2023}, visual transformers are just beginning to emerge in the visual domain. While the inherent flexibility of transformer architectures suggests potential for domain-specific customization through fine-tuning, the substantial computational investment required positions this as a relative weakness.

\textit{5. Cost (Data and Compute):}
The computational cost of deploying transformer-based models stands as a major hurdle. Illustrating the necessary scale of investment, \cite{rameshZeroShotTexttoImageGeneration2021} required 1024 Nvidia V100 GPUs for their project. Similarly, \cite{changMuseTextToImageGeneration2023} utilized 460 million text-image pairs, harnessing the power of 512 TPU-V4’s for over a week, to train their model. 
The substantial computational investments required underscore the practical challenges for applying transformer-based models in PDC, where such extensive resources may not be readily available. This is compounded by the relative scarcity of streamlined, well-understood methods for customizing these models to fit specific domain requirements.

\section{Application Recommendations}
\label{sec:recommend}
After analyzing each model family concerning the requirements for successful applicability in PDC, we aim to give a brief application recommendation in this chapter. We encourage the reader to carefully analyze their respective task and use our recommendations as a starting point to find the feasible generative model. We summarize our recommendations for each model family in \Cref{tab:recommend}.

\subsection{VAE's}
VAEs are particularly suited for tasks requiring the exploration of a range of design possibilities within a given design space. Their ability to interpolate smoothly between points in the learned distribution qualifies them for generating novel design variations. VAEs are well suited for early-stage ideation and design-space exploration. Ensuring that generated designs meet practical feasibility and technical plausibility is challenging with VAEs. They are best applied for low-fidelity representations of product design concepts, namely 2D-shapes \cite{yonekuraGeneratingVariousAirfoil2021} and structural images \cite{fanAdversarialLatentAutoencoder2023}. If the task requires high levels of visual details and fidelity, VAEs are not the right choice.
Efficiently applicable modalities for user-conditioning include class-labels, performance parameters, and simple visual inputs like sketches. More complex modalities like text and reference images usually require additional models for information embedding and injection.

VAEs can be trained efficiently on consumer-level or small industry-scale hardware. They require moderate amounts of data for domain-specific tasks. Compared to other generative models, the application of VAEs for domain-specific tasks poses a comparatively low skill threshold, as the architecture complexity is limited and there are vast amounts of literature and tutorials available.

\begin{figure}
    \centering
    \includegraphics[width=\linewidth]{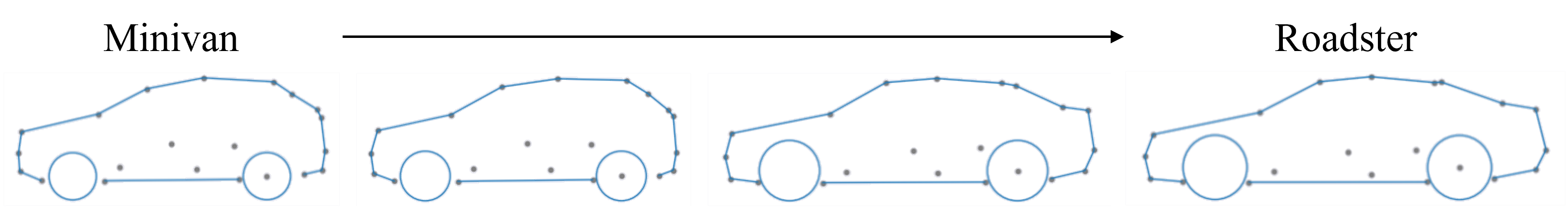}
    \caption{Interpolation between two classes in the learned CVAE.}
    \label{fig:VAE_Car}
\end{figure}

We implemented a conditional VAE architecture to generate simplified 2D-shapes of passenger vehicles (\Cref{fig:VAE_Car}). The generation is class-conditioned on six distinct categories. We use a dataset of 700 pairs of point-clouds and categorical features and train our model for half an hour on an NVIDIA RTX A4500 20GB. Details about our model architecture as well as other examples of VAE applied to product design tasks can be found in the appendix \cref{sec:Appendix} and in \Cref{fig:VAE_latent}. A summary of the recommendations is provided in the first column of \Cref{tab:recommend}.

\begin{table*}[h]
\footnotesize
\caption{Applicability recommendations in Product Design Conception.}
    \centering
    \begin{tabularx}{\linewidth}{ZZ|ZZ|ZZ|Z}
        \toprule
        & \multicolumn{1}{c|}{\textbf{VAEs}} & \multicolumn{2}{c|}{\textbf{GANs}} & \multicolumn{2}{c|}{\textbf{Diffusion Models}} & \multicolumn{1}{c}{\textbf{Transformers}} \\
        \midrule
        \textbf{Suitable Representation} & \textbf{2D-Shapes} & \textbf{2D Shapes} & \textbf{Natural Images} & \textbf{Low-fidelity Images} & \textbf{Natural Images} & \textbf{Natural Images}  \\
        \midrule
            \textbf{Generalization Capabilities} & Limitation to domain-specific tasks & Domain-specific tasks, interpolation in the learned distribution & Multi-category generation  & Domain-specific tasks, interpolation in the learned distribution & General purpose generation with domain knowledge & General purpose generation with domain knowledge \\
        \midrule
            \textbf{Possible Customization} & Training from scratch & Training from scratch & Training from scratch & Training from scratch	& Finetuning large-scale pretrained models &  Finetuning large-scale pretrained model\\
        \midrule
            \textbf{Required Data (Lower Threshold)} & Several hundred to few thousands of samples	& Several hundred to few thousands of samples & > 100.000 samples & 1.000-100.000 samples & Several hundred to few thousands of samples & 10.000-100.000 samples \\
        \midrule
            \textbf{Required Hardware} & Consumer-grade GPU & Consumer-grade to industry-level GPU (VRAM > 16GB) & Industry-level GPU cluster or cloud & Industry-level GPU (VRAM > 16GB) & Industry-level GPU (VRAM > 32GB) & Consumer-grade GPU or cloud \\
        \midrule
            \textbf{Conditioning Mechanisms} &	Class-labels, Performance Parameters, Visual Inputs (Sketch) & Performance Parameters, Visual Inputs (Sketch) & Category-labels, Performance Parameters &	Unconditioned, Text, Visual, Structural	& Text, Visual Inputs, Structural Inputs & Text\\
        \midrule   
            \textbf{Required Knowledge} & Basic knowledge & Experienced & Expert & Experienced-Expert & Experienced-Expert & Experienced-Expert\\
        \bottomrule
    \end{tabularx}
\label{tab:recommend}
\end{table*}

\subsection{GAN's}
In PDC, GANs are best utilized for generating 2D-shapes and images with low- to medium-fidelity content \cite{chenBezierGANAutomaticGeneration2021,chenMOPaDGANReparameterizingEngineering2021,ohDesignAutomationIntegrating2018}. Application examples for 2D-shapes and images are shown in \Cref{subsec:Gan_examp} and we summarize our recommendations in columns two and three in \Cref{tab:recommend}.

Due to adversarial training, they are suitable for generating representations with an increased level of detail compared to VAEs. The generalization capabilities depend on the size of the model and the dataset. While efficient for domain-limited applications, generalization capabilities that go beyond a small number of categories require extensive amounts of computational resources and are challenging to optimize due to training instabilities and hyperparameter-sensitivity. Successfully training GAN-based models requires advanced experience and expertise.

Suitable conditioning modalities for GAN-based models include performance parameters, category-labels, and visual inputs like sketches. Conditioning with text is certainly possible but increases the complexity of the overall architecture and requires additional models for text-embedding.

GAN-based methods require moderate amounts of training data depending on the fidelity of the target representation. The required amount of training data however rapidly rises with increasing generalization capabilities. They are adaptable to both consumer-level hardware for smaller models generating 2D-shapes and industry-scale GPUs for tasks involving image generation.

\subsection{Diffusion Models}
Diffusion models have emerged as a versatile and capable method for generating medium to high-fidelity images. Large-scale diffusion models, in particular, set a new standard for the quality and the level of detail of generated images. Small- to medium-sized diffusion models can be trained from scratch on mid-size industrial GPUs within a reasonable timeframe, making them accessible for domain-specific engineering and design applications \cite{fanNoiseSchedulingGenerating2023}. Our recommendations are summarized in in columns four and five of \Cref{tab:recommend}. We present an implementation of a DDIM \cite{songDenoisingDiffusionImplicit2022}, trained on an updated version of the BIKED dataset \cite{regenwetterBIKEDDatasetMachine2021} to generate image representations of bicycles in \Cref{subsec:Diff_examp} and \Cref{fig:ddim_biked} the appendix.

For projects requiring high-fidelity images, leveraging pre-trained general-purpose diffusion models is highly effective. These models can be fine-tuned with moderate amounts of data and computational resources to produce domain-specific content. Finetuning requires moderate amounts of computational resources and data. A mid-size industry-level GPU is usually sufficient. To an extent, large-scale models can even be used off-the-shelf without any training. Careful prompting (or other conditioning) can still yield valuable results, although the quality will be limited. Small-scale latent diffusion models can be trained from scratch on domain-specific data. However the image fidelity remains limited. High quality results can be achieved when these results act as domain-specific priors and are subsequently refined by a large-scale model like SDXL \cite{podellSDXLImprovingLatent2023} or FLUX \cite{flux2024}. Application examples showing results with off-the-shelf models, finetuned and refined models and different conditioning mechanisms are provided in \Cref{subsec:Diff_examp}.

In the realm of image-generation, diffusion models stand out for their exceptional flexibility in conditioning, capable of accommodating a vast range of inputs and adjustments. Conditioning modalities like text or visual inputs can be introduced using existing methods, significantly lowering the barrier for application and democratizing their usage. Modifying existing approaches to suit domain-specific needs, however, requires a high level of expertise.

\subsection{Transformers}
In PDC, transformer-based models are primarily feasible for tasks involving image and 3D-mesh generation. We provide some examples that show the capabilities of transformers with respect to the PDC-requirements in \Cref{subsec:Transformer_examp}. Our recommendations are summarized in the last column of \Cref{tab:recommend}.

For images, the level of detail and fidelity is surpassing  diffusion models. However, transformer-based models are infeasible to train from scratch for most domain-specific tasks. Therefore, utilizing pre-trained general-purpose models and finetuning them remains the only viable way for efficient application. The exploration of transformer-based models for visual tasks has only recently gained momentum. The tendency to produce representations that show increased real-world coherence compared to diffusion models and the predicted scalability make them a promising approach. 

However, the application of such models, especially in domain-specific product design, presents substantial challenges. The requirement for large amounts of computational resources and expert-level technical knowledge to effectively train and adapt these models poses barriers to accessibility and practical implementation. Mechanisms for custom conditioning without the necessity for finetuning or expensive adapters are yet to be developed. Additionally, many state-of-the-art approaches contain no publicly available code or model weights, making it difficult to verify the potentials for specific applications.

Consequently, while the potential for transformer-based visual models in product design is evident, realizing this potential within the constraints of domain-specific applications remains a complex endeavor.

\section{Conclusion}
\label{sec:Conlusion}
In this study, we explore the potential of Deep Generative Models (DGMs) in the context of their utility in PDC. We guide our analysis by the requirements that the PDC-process poses, specifically addressing open challenges in DGM-application.

VAE- and GAN-based DGMs are an appropriate choice for tasks with low to medium levels of detail in the visual representation and when the generalization requirements are limited to a few categories. With their versatility and robust capabilities, diffusion models stand out in generating high-quality, detailed visual content. These models are rapidly advancing, with ongoing enhancements in customization options and efficiency improvements that render them increasingly accessible for practical applications.

Transformer-based architectures are catching up and even surpassing diffusion models for image generation tasks. They can produce more coherent and contextually relevant outputs. Additionally, current limitations in generating more-complex and functional image representations are actively addressed by the scientific community.

Despite the potentials in employing DGMs to elevate traditional design processes by automating and enhancing creativity and efficiency, the integration of these technologies necessitates careful strategic planning and resource allocation. The successful adoption of DGMs hinges on the suitability of the use-case and the readiness of the environment in which they are deployed. Key factors such as data availability, computational resources, target representation adequacy, necessary conditioning mechanisms, expected accuracy levels, and the requisite expertise and knowledge must all be aligned for successful implementation.

In conclusion, while DGMs offer transformative potential for PDC, their full integration into the industry requires not just technological innovations but also a rethinking of how these tools can be made more accessible and practical for domain-specific, engineering- and design-oriented use. 

\section{Future Research Directions}
The field of DGMs is subject to enormous interest by the scientific community. This becomes especially evident for diffusion-and transformer-based models and the exponential growth of related publications (\Cref{fig:CV_Pubs}).

\textit{Conditioning and Customization:} Due to the rapid pace of technological development in DGMs, there remains a significant disconnect between improvements in general capabilities and the practical needs of designers and engineers. This is visualized by the discrepancy between the number of overall publications of DGMs and the number of publications concerning DGMs in engineering-design related subjects (\Cref{fig:CV_Pubs} and \Cref{fig:DGM_Pubs}).

\textit{Accessibility:} The accessibility-gap underscores the necessity for developments not just in capabilities, but in making these models more user-friendly and less resource intensive. For better applicability in engineering design and PDC, future research needs to focus on democratizing DGMs as advanced design tools across the industry. The challenge of accessibility includes not only model size but requiring less specialized knowledge to utilize, condition, modify and deploy DGMs.

Especially in large-scale models, accessibility is a critical challenge. While aiming to enhance the quality of outputs, these approaches generally demand more substantial computational resources, larger datasets, and greater expertise in customization and conditioning. This increased requirement not only makes them less accessible to a broader range of users but also raises concerns about the potential monopolization of their development and application. Such monopolization could mean that businesses with specific needs but without the capabilities to develop or customize these models might find themselves increasingly dependent on a few dominant players. This dependence could stifle innovation in sectors where companies cannot afford or manage the complexities involved in deploying advanced DGMs.

\textit{Datasets:} Research in image-and 3D-generation benefits from large-scale, generalized datasets like ImageNet, LAION, ShapeNet and ObjaVerse \cite{russakovskyImageNetLargeScale2015,schuhmannLAION5BOpenLargescale2022,changShapeNetInformationRich3D2015,deitkeObjaverseUniverseAnnotated2022,deitkeObjaverseXLUniverse10M2023}. The public availability of datasets meeting the specific requirements of product and engineering design is limited to small-scale, specific examples (e.g.: BIKED \cite{regenwetterBIKEDDatasetMachine2021}). Engineering-oriented domains require datasets that not only provide visual representations but also encompass performance parameters and detailed structural and geometric annotations to train models effectively. The development of such datasets is crucial.

\textit{Metrics:} Equally critical for the adoption of DGMs in product design is the development of appropriate metrics to assess model performance. Common metrics used in visual computing, such as FID \cite{heusel2018ganstrainedtimescaleupdate}, LPIPS \cite{zhangUnreasonableEffectivenessDeep2018}, and SSIM  \cite{wangImageQualityAssessment2004} fall short in evaluating the technical feasibility and real-world applicability of generated representations. For 3D-objects, current methods rely on metrics from the field of Computer Vision that define the similarity between point-clouds, like the Chamfer Distance (CD) \cite{wu2021densityawarechamferdistancecomprehensive}. There is no accurate possibility to assess overall realism, geometrical composition and feasibility of generated 3D-objects. 

The assessment of coherence and consistency with technical requirements remains a manual task in PDC-applications of DGMs and is heavily reliant on human evaluation. To bridge this gap, there is a pressing need for developing new metrics that can comprehensively evaluate the functional aspects of designs generated by DGMs.

\textit{Benchmarks:} Alongside large-scale open-source datasets and sophisticated metrics, DGMs benefit from the availability of benchmark challenges, which are crucial for capability evaluation and comparison. In image generation, there are numerous datasets used to capture the generative capabilities by calculating how well a given model can reproduce the initial data distribution (e.g.: CelebHQ \cite{karrasPROGRESSIVEGROWINGGANS2018}, FFHQ \cite{karrasAnalyzingImprovingImage2020}, LSUN \cite{yuLSUNConstructionLargescale2016} and MS COCO \cite{linMicrosoftCOCOCommon2015}). In 3D-generation, Google Scanned Objects (GSO) is widespread for benchmarking \cite{downsGoogleScannedObjects2022}.

To improve DGMs applied in engineering and design related domains, there is a need for the initialization of benchmark problems. These should be aligned with the methodology of existing benchmarks in image and 3D generation to make them more accessible. They should allow for the evaluation of DGM-capabilities in the requirements for PDC-application like geometric consistency and alignment with input conditions, technical feasibility and performance evaluation. 


\bibliographystyle{asmejour}   

\bibliography{main} 



\clearpage
\appendix
\selectlanguage{english}
\section*{Appendix}
\label{sec:Appendix}

\section{Publication Quantities for DGM-Families}

\begin{figure}[h]
\centering
    \includegraphics[width =0.8\linewidth, height = 4cm]{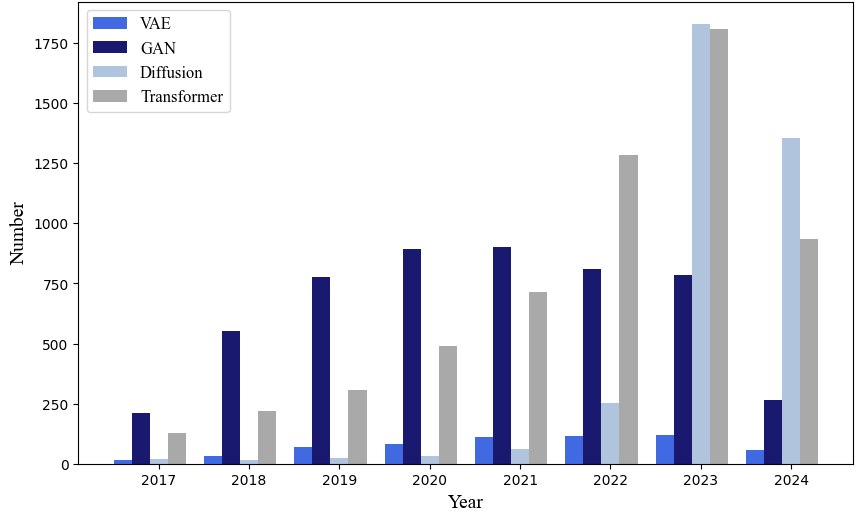}
    \caption{Quantity of papers on ArXiv in the Computer Vision category (cs.CV), mentioning the model type in their abstract}
    \label{fig:CV_Pubs} 
\end{figure}

\begin{figure}[h]
\centering
    \includegraphics[width =0.8\linewidth, height = 4cm]{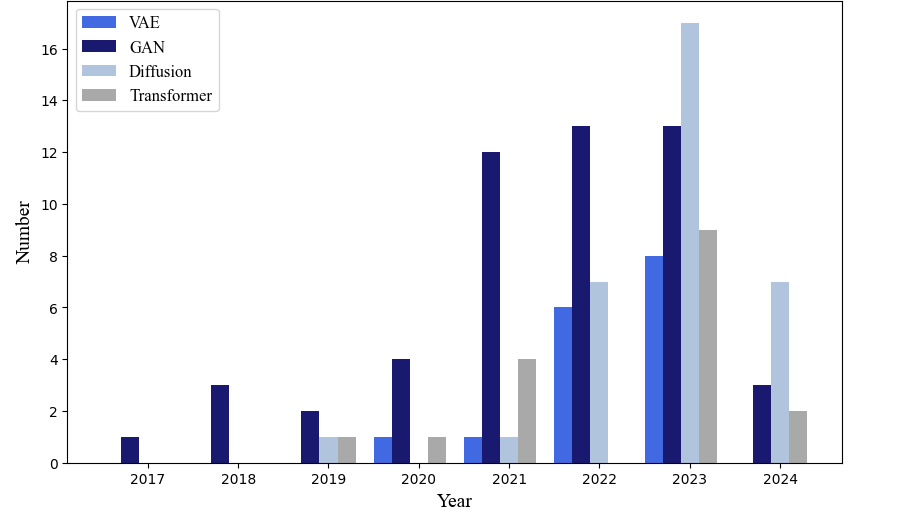}
    \caption{Quantity of papers on ArX iv in the Machine Learning Category (cs.LG ), mentioning the model type in their abstract.}
    \label{fig:DGM_Pubs} 
\end{figure}
\FloatBarrier

\section{DGM-Applications in PDC}
\subsection{Examples of VAE-based Generation}
\label{subsec:VAE_examps}

\textit{Generation of 2D Shapes:} In PDC, VAEs are beneficial for low-fidelity representations of shapes with limited amounts of data and compute. This is shown by our implementation of a cVAE to generate 2D shapes of vehicles. The point-cloud consists of 21 structured geometric reference points.

\begin{figure*}[h]
    \centering
    \includegraphics[width=0.9\textwidth]{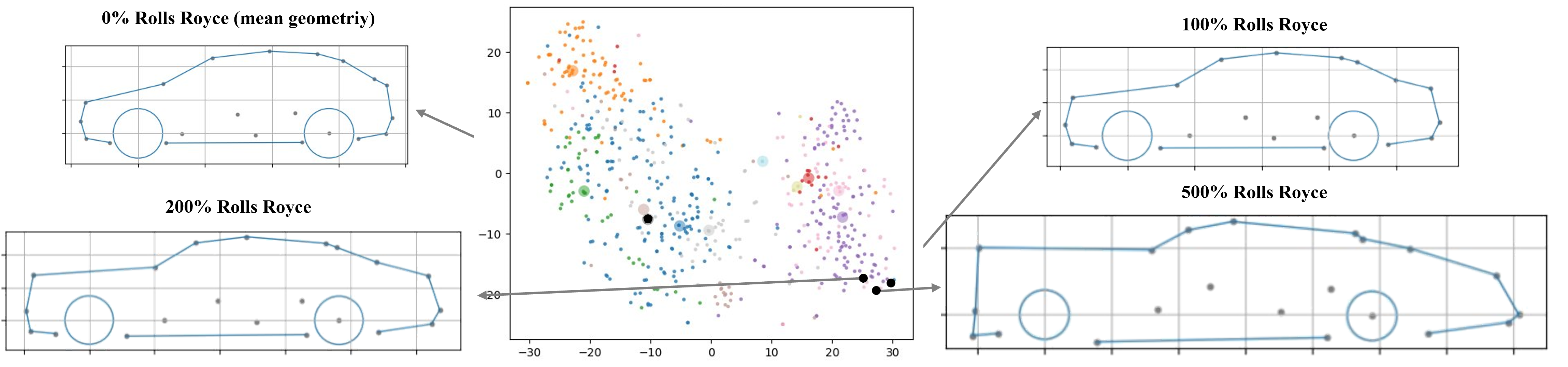}
    \caption{Attribute extrapolation in the latent space of the cVAE. Attribute from class "Manufacturer"}
    \label{fig:VAE_latent}
\end{figure*}

The categorical conditions are one-hot encoded. Training, with a batch size of 64, was conducted over 5,500 epochs using an Adam optimizer with a learning rate of 0.0002. The ratio of KL-divergence loss and reconstruction loss MSE) in the loss function is 0.4/0.6. The simple architecture is capable of interpolations between different conditional categories (\Cref{fig:VAE_Car}) as well as extrapolating a condition (\Cref{fig:VAE_latent}).

\subsection{Examples of GAN-based Generation}
\label{subsec:Gan_examp}
\begin{figure*}[h]
    \centering
    \includegraphics[width=0.9\textwidth]{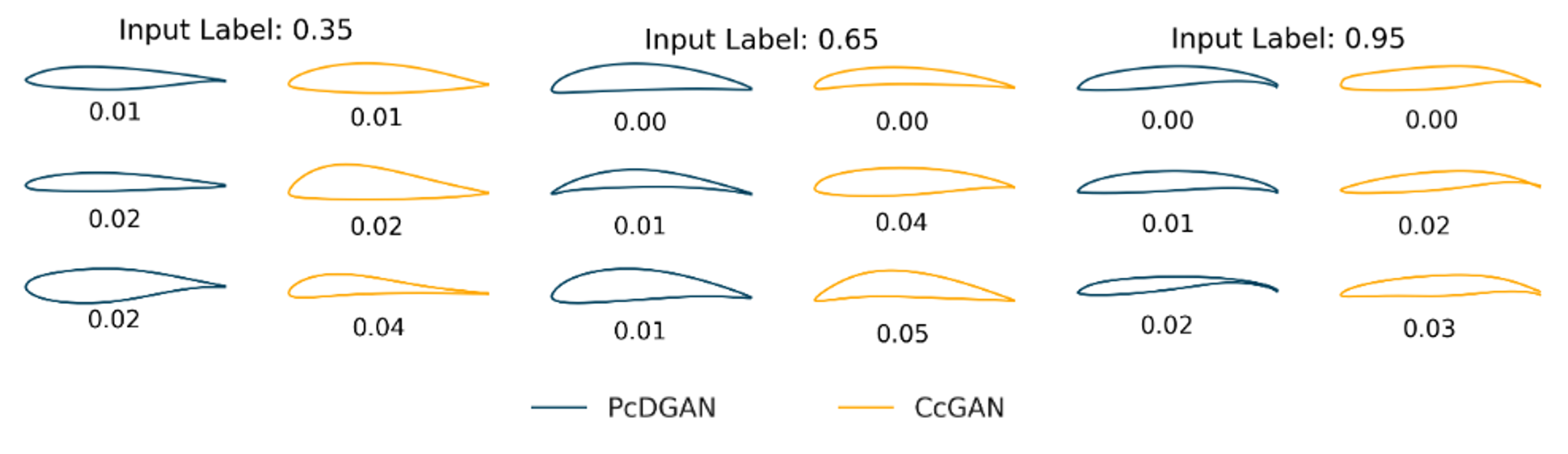}
    \caption{Generated airfoil profiles by PcDGAN \cite{nobariPcDGANContinuousConditional2021}and CcGAN \cite{dingCCGANCONTINUOUSCONDITIONAL2021}, conditioned on the Lift -/Drag
    Ratio. The label error is shown below every airfoil. Examples taken from the PcDGAN publication \cite{nobariPcDGANContinuousConditional2021}.}
    \label{fig:airfoil_gan}
\end{figure*}

\textit{Generation of 2D-Representations:} GAN-based approaches are an efficient solution for low-fidelity representations of technical concepts, as they can be trained to adhere to technical conditions with moderate amounts of data. \Cref{fig:airfoil_gan} depicts results of two approaches generating simple airfoil shapes, conditioned on the lift-/drag-ratio as the performance parameter \cite{nobariPcDGANContinuousConditional2021}.

\begin{figure*}[h]
    \centering
    \includegraphics[width=0.8\textwidth]{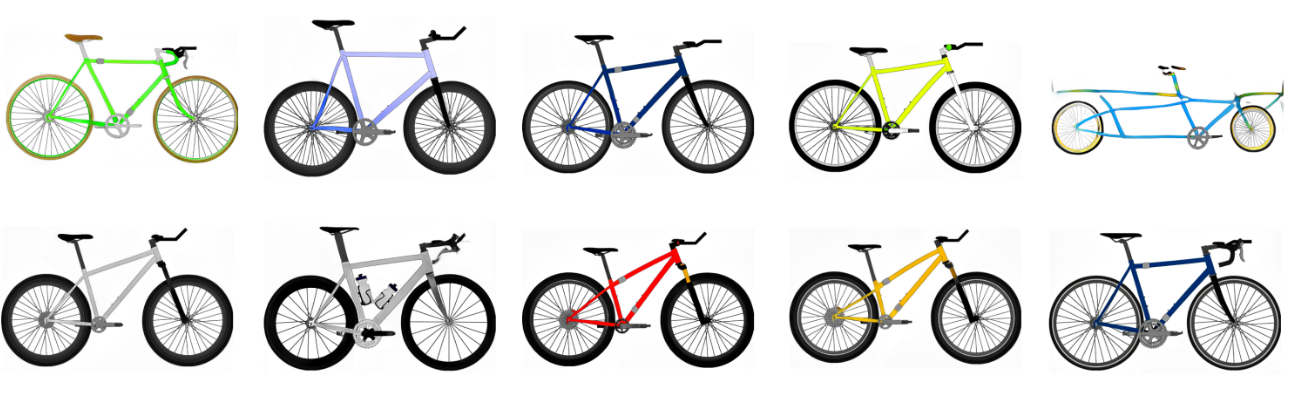}
    \caption{Generated samples by CreativeGAN \cite{nobariCreativeGANEditingGenerative2021} trained on the BIKED dataset \cite{regenwetterBIKEDDatasetComputational2021}. Image taken from the CreativeGAN publication \cite{nobariCreativeGANEditingGenerative2021}}
    \label{fig:creative_gan_biked}
\end{figure*}

They can also be used for image representations of technical concepts with higher fidelity and more geometric details. \Cref{fig:creative_gan_biked} visualizes generated samples by CreativeGAN \cite{nobariCreativeGANEditingGenerative2021}, a model focusing on producing results with a high degree of novelty. Training is based on a medium size dataset with 4775 bicycle images. The example shows that realistic and geometrically consistent concepts can be generated efficiently for limited domains.

\subsection{Examples of Diffusion-based Generation}
\hfill
\label{subsec:Diff_examp}
\textit{DDIM for Generation of Structural Bicycle Images:} Diffusion models are a viable choice for a variety of tasks in PDC. For medium-fidelity image representations they can be trained from scratch. The diffusion model we use for generating black-and-white images of bicycle geometries follows the DDIM-architecture \cite{songDenoisingDiffusionImplicit2022}. We use the model structure and hyperparameters proposed in Ref. \cite{fanNoiseSchedulingGenerating2023}. This architecture includes a U-Net with six feature map resolutions, ranging from 256x256 to 4x4. Each upscaling and downscaling layer incorporate one residual block, and an attention layer is featured at the 16x16 resolution. 1,000 timesteps and a linear noising schedule, where $\beta$ starts at 1e-4 and increases to 0.02, are employed.

\begin{figure*}[h]
    \centering
    \includegraphics[width=0.9\linewidth]{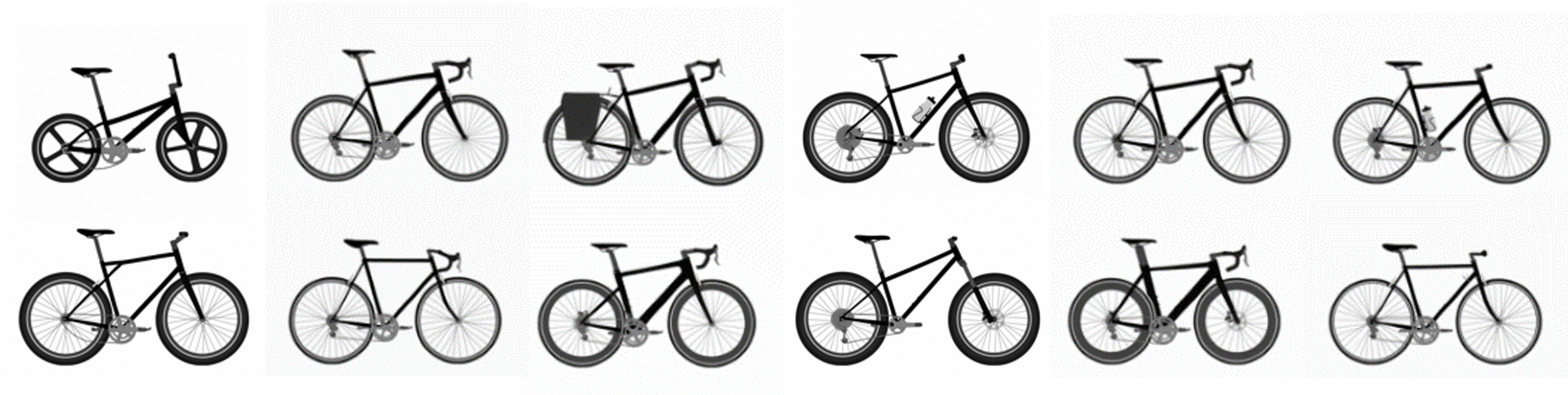}
    \caption{Images of bicycle geometries generated by the DDIM.}
    \label{fig:ddim_biked}
\end{figure*}
The dataset, randomly sampled examples are shown in \Cref{fig:ddim_biked} comprises 4,300 images of bicycle geometries, sized at 256x256 pixels. Prior to training, we curated the dataset by removing unrealistic and faulty samples, standardizing, and maximizing the scaling of the geometries within the images. The model was trained for 291000 epochs with a batch-size of 16, using an Adam optimizer with a learning rate of 0.00005.

\textit{Generation of Product Images with Self-Trained Latent Diffusion Models:}
Large-scale diffusion models for high-quality natural image synthesis are trained on hundreds of millions of images. They are infeasible to train from scratch for domain specific applications. However, latent diffusion models for special, limited applications can be trained from scratch, only requiring moderate amounts of compute. We train a Latent Diffusion Model (LDM) based on the PlantLDM \cite{FisherPlantLDM2022}, which is a simplified unconditional implementation of Stable Diffusion \cite{rombachHighResolutionImageSynthesis2022}. In our experiments, we use two image datasets. We use GeoBIKED \cite{muellerGeoBiked2024}, which provides image representations of the bicycles in addition to categorical and numerical features. We also use the DVM-Car dataset \cite{Huangdvmcar2022}. The quality-checked subset of DVM contains 67k front view images of passenger vehicles. For annotations, we use 14 categorical and 10 numerical features. 

The categorical conditioning information is embedded and concatenated to the latent image representations of the U-Net. In the U-Net, we increase the starting channel dimension to 64 for the bicycle data and to 128 for the car data. The U-Net levels are set to 4 and the number of attention heads is increased to 8 for both models. We use the pretrained Stable Diffusion 2.1 VAE for encoding the images. For the model trained on GeoBiked, we employ a batch size of 32. The DVM-Car model is trained with a larger batch size of 128. The learning rate is $1e{-4}$ for both. We emphasize that large-scale, state-of-the-art LDMs achieve notably higher image fidelity. Given the relatively small size of the dataset, the lack of specific optimization and the moderate computational cost (approx. 30 hours on a single A6000), these models achieve moderate image fidelity, but show in principle that small scale diffusion models can be trained from scratch for use-case specific applications. 
\begin{figure*}
    \centering
    \includegraphics[width=\linewidth]{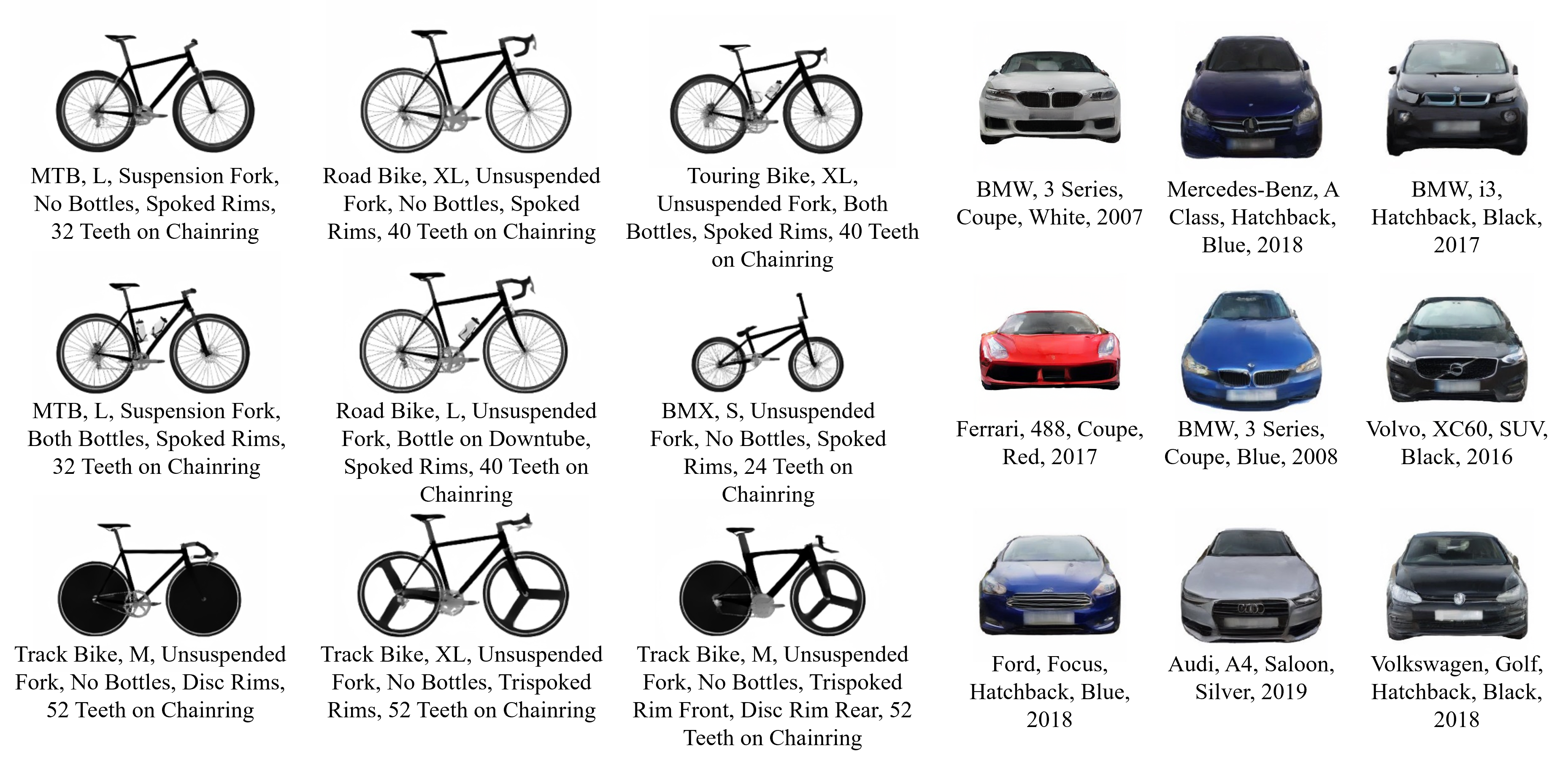}
    \caption{Qualitative examples of product images generated with our trained LDM. Left: LDM trained on GeoBiked \cite{muellerGeoBiked2024}. Right: LDM trained on DVM-Cars \cite{Huangdvmcar2022}.}
    \label{fig:ldm_examples}
\end{figure*}

\begin{figure*}
    \centering
    \includegraphics[width=0.9\textwidth]{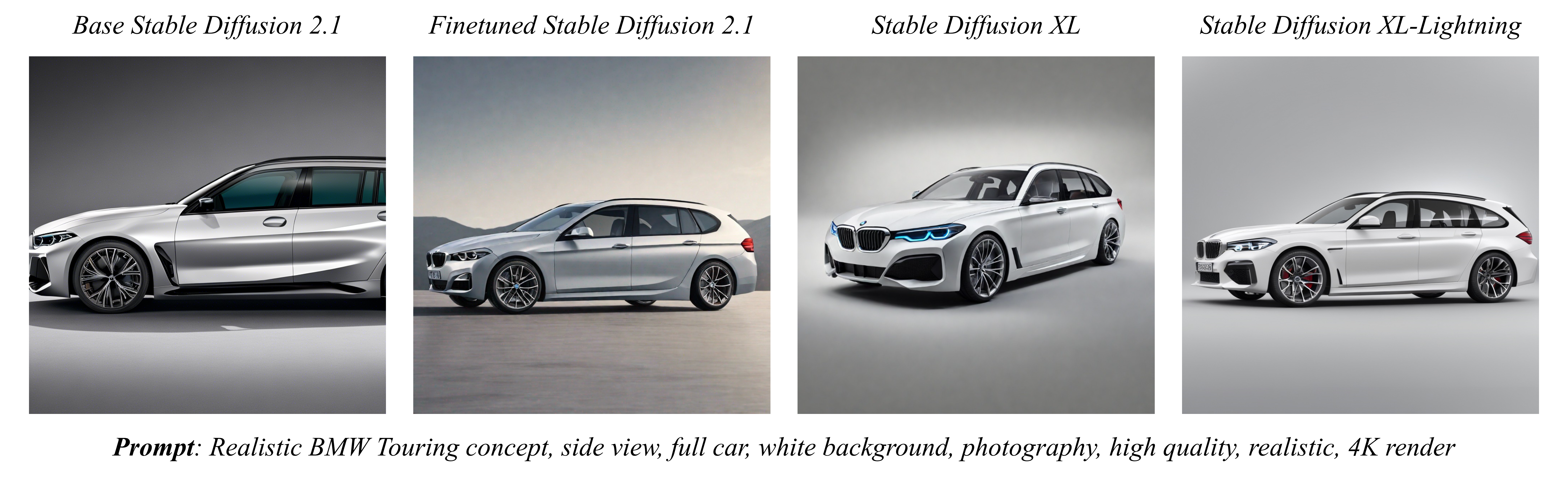}
    \caption{Comparison of different versions of the Stable Diffusion model for image generation. 50 sampling steps with DDIM sampler were used for all generations. Models: Stable Diffusion \cite{rombachHighResolutionImageSynthesis2022}, Stable Diffusion XL \cite{podellSDXLImprovingLatent2023} and SDXL-Lightning \cite{linSDXLLightningProgressiveAdversarial2024}.}
    \label{fig:sd_bmw}
\end{figure*}

\textit{Generation and Conditioning of Product Concept Images with Stable Diffusion:}
For natural image representations of product concepts that contain more details and higher levels of realism, diffusion models are the go-to-choice, offering high visual quality and adequate real-world coherence while providing a wide variety of conditioning mechanisms. \Cref{fig:sd_bmw} compares the results of different versions of Stable Diffusion implementations conditioned on the same input prompt.

While the base model fails to capture the required spatial and structural information (“side view” and “full car”) and the technical feasibility is limited (e.g.: the wheel) the finetuned model shows a significant improvement in those aspects. Stable Diffusion XL achieves similar results without any finetuning on domain-specific data (about 2000 images of BMW vehicles in our case).

\begin{figure*}
    \centering
    \includegraphics[width=0.9\textwidth]{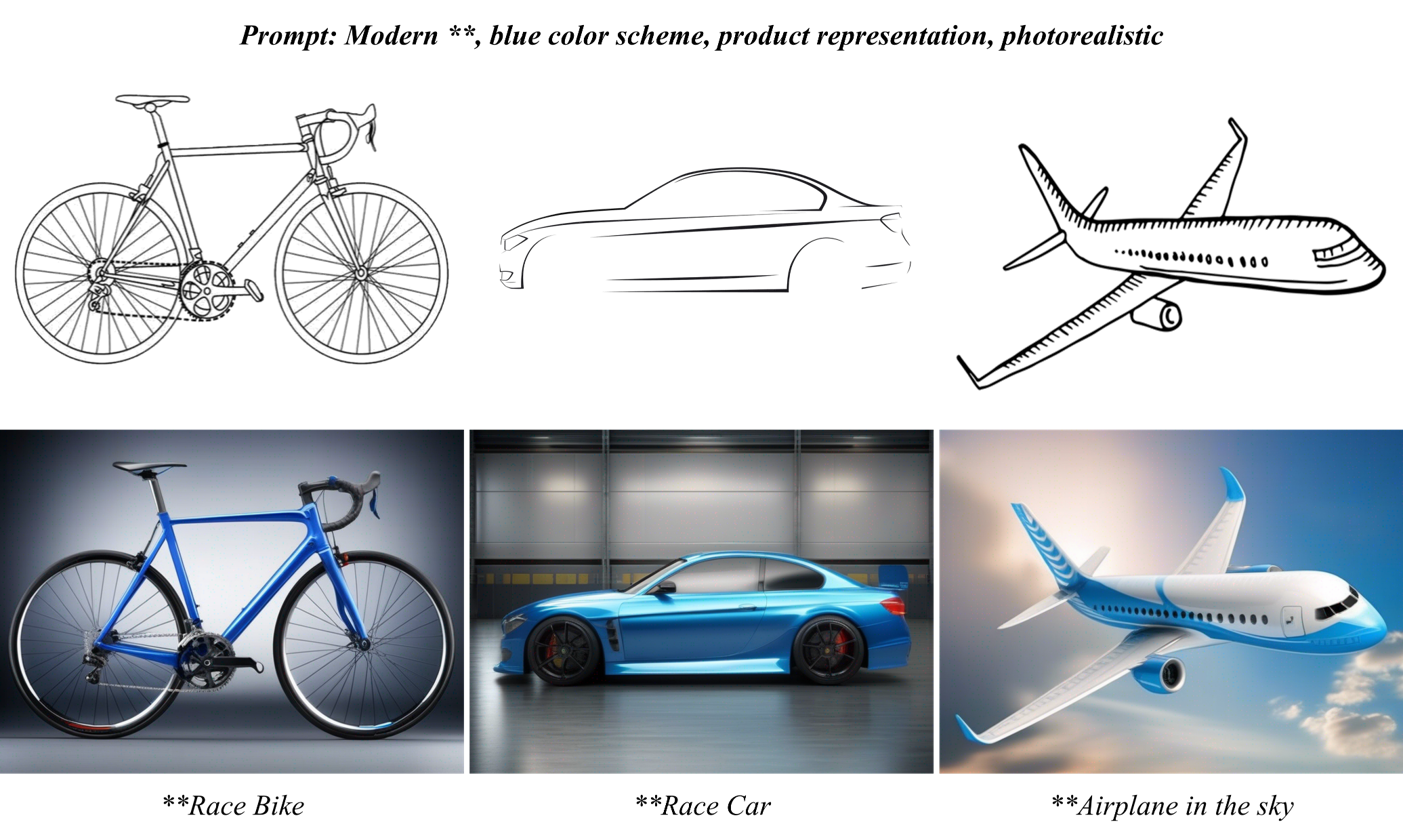}
    \caption{Image representations of product concepts generated by text-conditioned Stable Diffusion and a sketch-conditioned ControlNet adapter \cite{zhangAddingConditionalControl2023}.}
    \label{fig:controlnet_example}
\end{figure*}

\begin{figure*}
    \centering
    \includegraphics[width=0.9\textwidth]{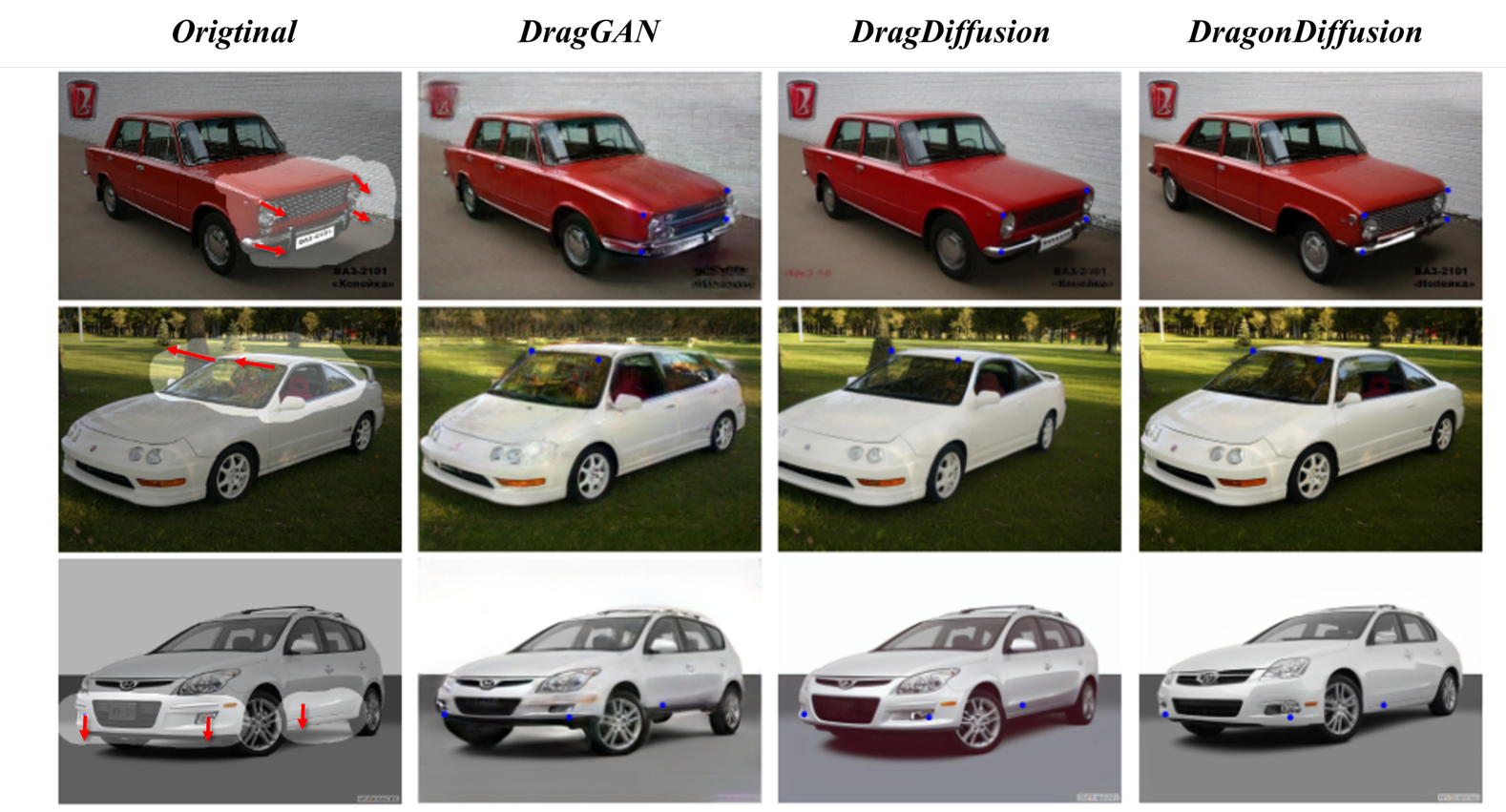}
    \caption{Modification of technical product images through point-dragging approaches. Image taken from DragonDiffusion publication. Models: DragGAN \cite{panDragYourGAN2023}, DragDiffusion \cite{shiDragDiffusionHarnessingDiffusion2023}, DragonDiffusion \cite{mouDragonDiffusionEnablingDragstyle2023}.}
    \label{fig:dragdiff_example}
\end{figure*}

\Cref{fig:controlnet_example} and \Cref{fig:dragdiff_example} show examples of conditioning mechanisms for diffusion-based image generation other than text-prompts. Both examples use the original implementations by the authors and have not been finetuned or adapted. The former example, ControlNet \cite{zhangAddingConditionalControl2023}, allows for a sketch as additional input, qualifying it for early-stage design processes. 
The latter image is taken from DragonDiffusion \cite{mouDragonDiffusionEnablingDragstyle2023} and compares different implementations of “dragging”-mechanisms where the user can select multiple seed- and target-points in an image and thereby change the structural composition of the product representation. This direct and intuitive control over geometrical characteristics of products bears potential for gradual design modifications and iterative adjustments of the product concept.

\subsection{Examples of Transformer-based Generation}
\label{subsec:Transformer_examp}
\begin{figure*}[h]
    \centering
    \includegraphics[width=0.9\textwidth]{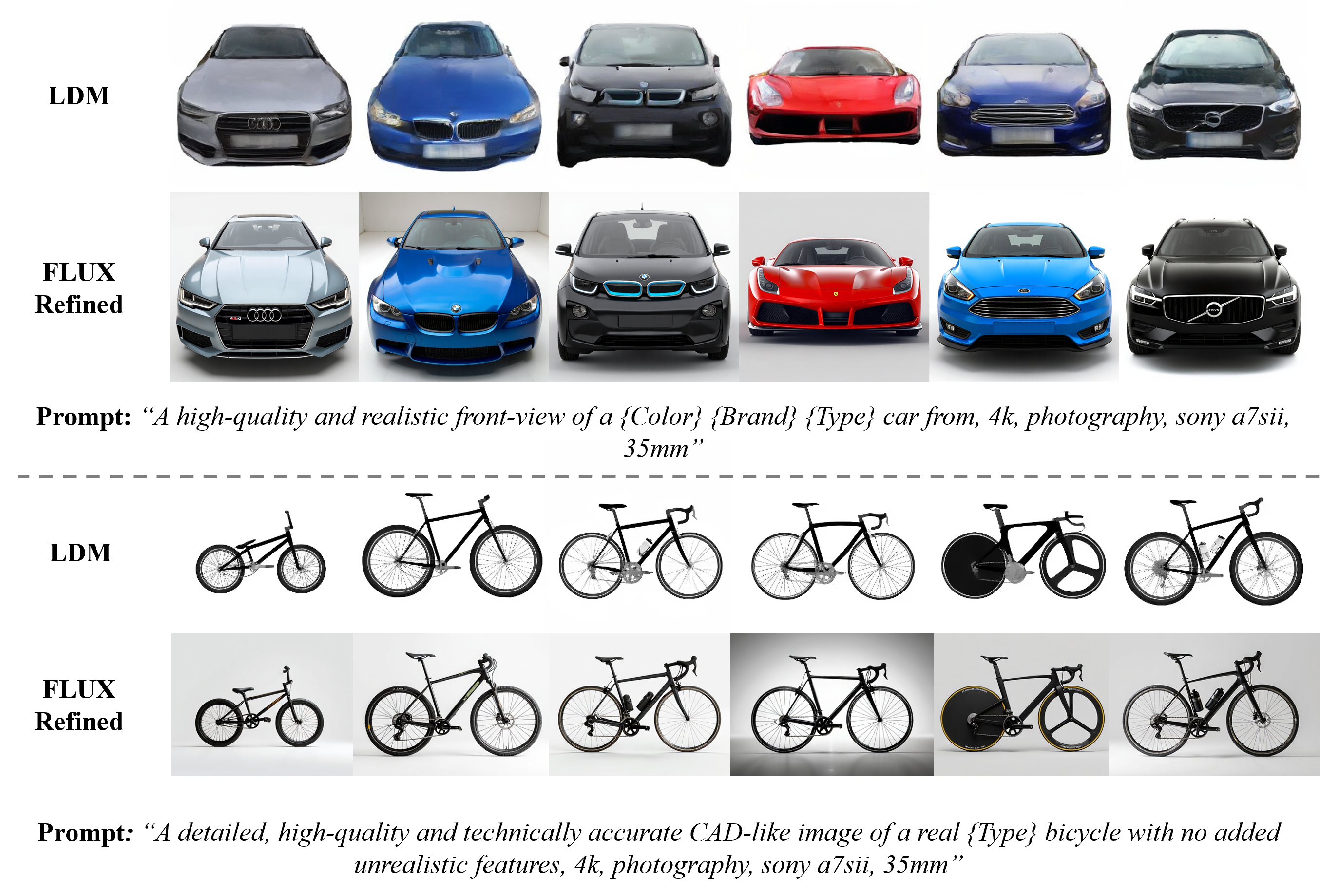}
    \caption{Images generated with our trained Latent Diffusion Model and refined with the transformer-based model FLUX \cite{flux2024, esserScalingRectifiedFlow2024}. (Top: LDM trained on DVM-Cars \cite{Huangdvmcar2022}, Bottom: LDM trained on GeoBiked \cite{muellerGeoBiked2024}.}
    \label{fig:FLUX_examples}
\end{figure*}

Transformer-based approaches for the generation of visual content promise increased control over the structural, spatial and contextual characteristics. This is depicted in state-of-the-art models for image generation \cite{flux2024, esserScalingRectifiedFlow2024}. Their applicability in PDC is currently limited and requires substantial resources and expert-knowledge. There are only few large-scale pretrained models publicly available \cite{flux2024}. Conditioning the generation requires adapters that are finetuned for the specific modalities, some are provided with FLUX.
\Cref{fig:FLUX_examples} shows some example images where control over these characteristics is achieved through a combination of providing a low-resolution reference image, generated by the much smaller LDM described in \Cref{subsec:Diff_examp} and additional structural guidance and text-conditioning.

\end{document}